\documentclass{article}

\usepackage[main, final]{neurips_2026}

\usepackage[utf8]{inputenc} 
\usepackage[T1]{fontenc}    
\usepackage{hyperref}       
\usepackage{url}            
\usepackage{booktabs}       
\usepackage{amsfonts}       
\usepackage{nicefrac}       
\usepackage{microtype}      
\usepackage{xcolor}         
\usepackage{amsmath}

\usepackage{comment}
\usepackage{subcaption}
\usepackage{graphicx}

\usepackage{tikz}
\usetikzlibrary{arrows.meta, positioning, fit}

\usepackage{algorithm}
\usepackage{algpseudocode}
\usepackage{etoolbox}
\usepackage{multirow}
\usepackage{makecell}
\usepackage{verbatim}
\usepackage{subcaption}
\usepackage{makecell}
\usepackage{wrapfig}
\usepackage{caption}

\title{Depth through Recurrence: Towards Ultra-Efficient On-Device ASR}

\let\SUP\textsuperscript
\author{%
    Chen Feng\SUP{1},
    Tianyi Xu\SUP{2}\thanks{McGill University. Work conducted during an internship at Qualcomm AI Research.},
    Ramchalam Kinattinkara Ramakrishnan\SUP{3},\\
    Yicheng Lin\SUP{4},
    Shaojie Zhuo\SUP{5},
    Zhaocong Yuan\SUP{6}, 
    Chenzheng Su\SUP{7}, 
    Xiaopeng Zhang\SUP{8}\\
    \\
    Qualcomm AI Research \thanks{Qualcomm AI Research is an initiative of Qualcomm Technologies, Inc.} \\
    \texttt{\{\SUP{1}chenf,\SUP{3}rkinatti,\SUP{4}yichengl,\SUP{5}shaojiez,\SUP{6}zhaocong,\SUP{7}chenzhen,\SUP{8}xiaopeng\}}\\
    \texttt{@qti.qualcomm.com}\\
    \texttt{\SUP{2}tianyi.xu2@mail.mcgill.ca}
}

\begin{document}

\maketitle

\begin{abstract}
Modern automatic speech recognition (ASR) systems have achieved remarkable accuracy by scaling model depth and capacity, but at the cost of increased memory, energy, and computational demands. On edge devices where ASR is often most needed, such as watches and glasses, deploying such large models is infeasible due to extremely constrained resources. This raises a fundamental question: can we achieve high representational power in deep ASR models without scaling up parameterization? In this work, we revisit the role of depth and identify layer-wise representational dynamics, in which most layers learn functionally similar transformations. Motivated by this insight, we propose a block-recurrent ASR architecture that replaces parameterized depth with a small set of recurrent blocks, each consisting of weight-shared layers, thereby preserving effective depth while drastically reducing model size. Through representation-guided grouping and knowledge distillation, block-recurrent models retain the accuracy of large reference models while using an order of magnitude fewer parameters. Across Open ASR Leaderboard benchmarks, a model with only two shared blocks recovers $97.4\%$ of the accuracy of large models on average. The compact backbone further enables efficient on-device personalization through lightweight MoE-LoRA adaptation, allowing user-specific models to match or surpass the accuracy of general-purpose ASR systems with significantly lower inference cost. By reusing weights across recurrent steps, the proposed architecture reduces memory-access overhead and inference energy consumption. Together, these results show that high-quality ASR can be achieved without large parameterization, and that structured recurrence provides a principled path towards ultra-compressed, energy-efficient, and personalized speech recognition on the edge.

\end{abstract}

\section{Introduction}

\label{sec:introduction}

Speech is emerging as the primary interface for next-generation human-machine interaction, catalyzed by advances in conversational AI and the proliferation of hands-free, screen-less devices that demand low-latency interaction \cite{speechgpt}. As computing shifts toward ambient systems, wearable devices, and physically situated AI, reliable Automatic Speech Recognition (ASR) becomes essential for enabling continuous multi-turn dialogue and hands-free control \cite{whisper}. Recent breakthroughs in ASR are largely driven by deep neural architectures such as Transformers \cite{transformerpp} and Conformers \cite{conformer, fastconformer}, which scales effectively with data, model size and capacity. As a result, state-of-the-art (SOTA) ASR models often comprise hundreds of millions to billions of parameters \cite{owls2025scaling}, delivering high accuracy across diverse speakers, acoustic conditions, and languages.

Despite their high accuracy, a significant gap remains between large ASR models and the physical constraints of the edge devices they are intended to power. While SOTA models thrive in cloud environments with abundant compute resources, deploying them on ultra-low-resource edge platforms, such as wearable, mobile and edge SoCs, laptops, and onboard robotic sensors, is fundamentally limited by extreme hardware constraints. Unlike server-class GPUs, these devices are typically restricted to tightly bounded memory footprints, low compute frequencies, and strict power budgets to maintain viable battery life. Moreover, they are further constrained by real-time latency requirements and limited memory bandwidth, which make it difficult to sustain the high-throughput inference by on-device ASR. 

Bridging the gap between the demands of on-device ASR system and these rigid hardware boundaries remains an open and critical challenge. On-device ASR offers compelling advantages over cloud-dependent systems \cite{lowbit}, including always-on operation, ultra-low latency, enhanced privacy, and the reliable offline autonomy required by physical AI systems. Therefore, adapting high-accuracy ASR to fit within the strict memory, compute, and power envelopes of on-device platforms is essential for enabling scalable, localized human-machine interaction.



Existing approaches to enabling on-device ASR primarily target post-training model compression, utilizing techniques such as quantization \cite{edgeasr, lowbit}, pruning \cite{pruning}, and low-rank approximation \cite{lowrankasr, molora2024}. While these methods effectively reduce the memory footprint and computational cost, scaling down billion-parameter SOTA models via compression alone often leaves them too large for the megabyte-scale confinements of embedded processors and IoT systems \cite{openasr}. Furthermore, pushing these techniques too aggressively inevitably leads to noticeable accuracy degradation, as deep acoustic encoders lose the capacity to reliably process complex audio signals \cite{acoustic}. Crucially, these post-training methods are orthogonal to the underlying architectural efficiency of the model itself. To break through the current Pareto frontier of size versus accuracy, we must address structural redundancies inherent in network design, a strategy that serves as a highly effective complement to existing compression techniques.


Recent work on inverse depth scaling laws \cite{inverse, universalweightsubspace, resnetensemble} highlights a fundamental inefficiency in how depth is currently utilized in modern architectures. In conventional designs, each layer is parameterized independently, leading to a linear growth in model size with depth. Prior analyses in abstract reasoning \cite{trn, hrn} and computer vision \cite{vits} show that transformations across layers are often highly correlated, indicating significant redundancy in these independently parameterized layers. Building on this observation, prior work \cite{ALBERT, UniTrans, RRTrans} adopts uniform parameter sharing as an architectural prior, where parameters are shared statically across layers. These studies further demonstrate that compact, block-recurrent networks can recover much of the functional expressivity of massive networks while significantly reducing the total parameter count. This conceptual shift raises a fundamental and largely unexplored question for ASR systems: \emph{can the effective depth and high accuracy of modern ASR models be preserved by exploiting their underlying representation dynamics-specifically through block-recurrent weight sharing-thereby enabling compact yet highly accurate models for on-device inference?}


In this work, we conduct a systematic representational analysis of large-scale ASR models. Using Centered Kernel Alignment (CKA) \cite{cka}, we uncover distinct, structured layer-wise similarities within deep encoders. Specifically, we observe that adjacent layers frequently learn functionally similar transformations, acting as an implicit ensemble of iterative refinement steps. This confirms that while deeper networks achieve higher recognition accuracy, the auxiliary layers within specific encoder regions primarily contribute incremental refinement rather than generating qualitatively new representational stages.



\begin{figure}[!t]
  \centering
  \vspace{-5pt}
  \includegraphics[width=0.8\linewidth]{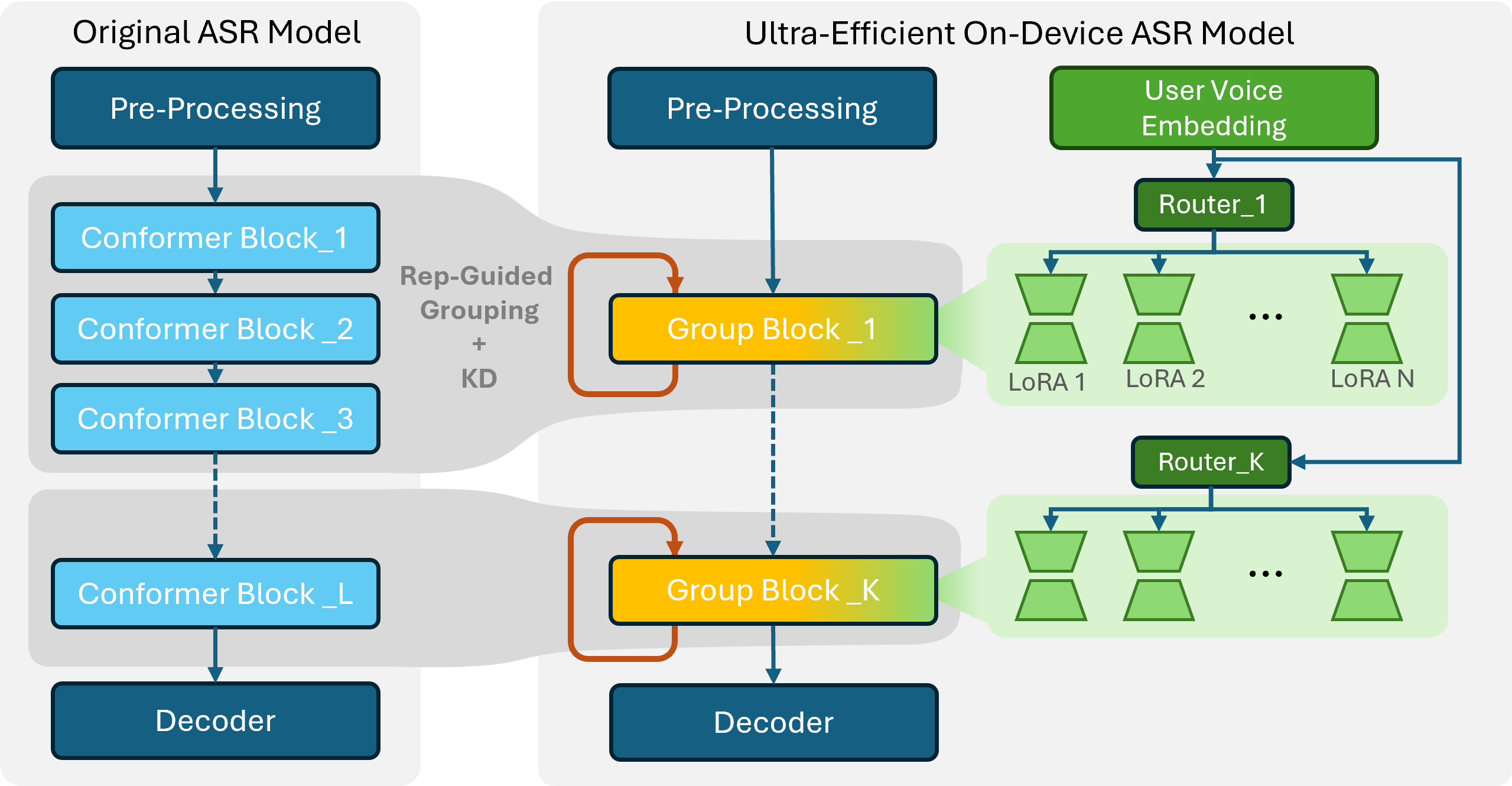}
  \caption{Overview of the proposed block-recurrent ASR framework. A large-scale ASR composed of $L$ encoder blocks is analyzed using CKA to identify representational similar layers. Guided by this analysis, an edge-scale ASR model is constructed with block-recurrence, and trained through knowledge distillation. The resulting compact backbone further enables efficient on-device personalization through MoE-LoRA adapters conditioned on user embeddings.}
  \label{fig:block_recurrent_asr}
  \vspace{-8pt}
\end{figure}


Motivated by these insights, we propose a scalable \textbf{block-recurrent architecture for ASR}, explicitly designed to factorize deep acoustic encoder into a compact set of recurrently applied blocks. Rather than dedicating unique parameters to every block, we partition the deep "teacher" model into groups of blocks based on their representational similarity. Each group is then replaced by a single, shared block that is applied recurrently, recovering the effective depth and computational structure of the original model. Using knowledge distillation, the resulting block-recurrent models shrink the footprint to mere tens of millions of parameters, often achieving an order-of-magnitude reduction compared to their teachers, while preserving high recognition accuracy.


We evaluate the proposed approach on two representative ASR models, covering both streaming and non-streaming settings, across eight benchmarks following the standard Open ASR Leaderboard protocols. Our results demonstrate that block-recurrent models achieve highly competitive Word Error Rates (WER) at a fraction of the original size, enabling practical on-device deployment. In particular, for the STT-480ms streaming model, a compressed encoder parameterized by only two shared Conformer blocks recovers $97.4\%$ of the reference model's accuracy, despite an order-of-magnitude reduction in parameters. Furthermore, the block-recurrent design inherently optimizes on-device execution: even with comparable FLOP budgets, the repeated reuse of weights dramatically improves memory cache locality, yielding higher practical efficiency and lower end-to-end latency.


Beyond the benefits of compression, the compact block-recurrent architecture provides an ideal foundation for on-device personalization. We introduce a lightweight adaptation framework utilizing Mixture-of-Experts Low-Rank Adaptation (MoE-LoRA). In this setup, user voice embeddings are extracted from a brief enrollment audio sample, and used to dynamically route and combine specialized LoRA adapters. This design enables the on-device model to seamlessly adapt to individual speakers and usage contexts, allowing personalized models to reliably match or even surpass the accuracy of general-purpose ASR systems with significantly lower inference cost.




\textbf{Contributions.} (1) We present a systematic empirical analysis of multiple state-of-the-art ASR models, revealing strong structured layer-wise representational dynamics in deep speech encoders. (2) We propose a block-recurrent ASR architecture and training recipe, in which encoder depth is factorized into a small set of CKA-guided shared blocks applied recurrently, yielding compact yet expressive models for on-device deployment. (3) We provide extensive experimental validation on eight benchmarks using the Open ASR Leaderboard protocol, demonstrating that block-recurrent models achieve near-reference recognition accuracy with an order-of-magnitude fewer parameters. On-device memory and energy efficiency are systematically profiled across diverse hardware platforms and deployment settings. (4) We further introduce an on-device personalization extension based on lightweight, user embedding conditioned MoE-LoRA adaptation, enabling efficient and privacy-preserving user-specific refinement of block-recurrent models. 

Overall, our work demonstrates that high-accuracy ASR can be achieved without large parameterization, and that structured recurrence provides a principled path towards ultra-compressed, energy-efficient, high-quality, and personalized speech recognition on the edge.

\section{Layer-wise Representational Dynamics}
\label{sec:inverse_depth_scaling}

To investigate the extent of layer-wise redundancy within state-of-the-art ASR architectures, we analyze four models: STT-480ms ($115$M)\cite{stt}, Parakeet-TDT-0.6B-v2 ($618$M)\cite{parakeet}, Canary-1B-Flash ($883$M)\cite{canaryqwen} and Granite-4.0-1B-Speech ($2.3$B)\cite{granite}. These models differ in architecture and encoder depth, while achieving top-tier ASR accuracy on the leaderboard \cite{openasr}. Since modern ASR systems are predominantly encoder-heavy in both parameter count and computational cost, we focus our analysis on restructuring the encoder to maximize representational efficiency. Linear \emph{Centered Kernel Alignment (CKA)} \cite{cka} is used to quantify layer-wise representational similarity across depth by comparing encoder activations extracted from a set of calibration speech samples.

\begin{figure}[h]
  \centering
  \vspace{-6pt}
  \includegraphics[width=1.0\linewidth, height=0.15\textheight,
  keepaspectratio]{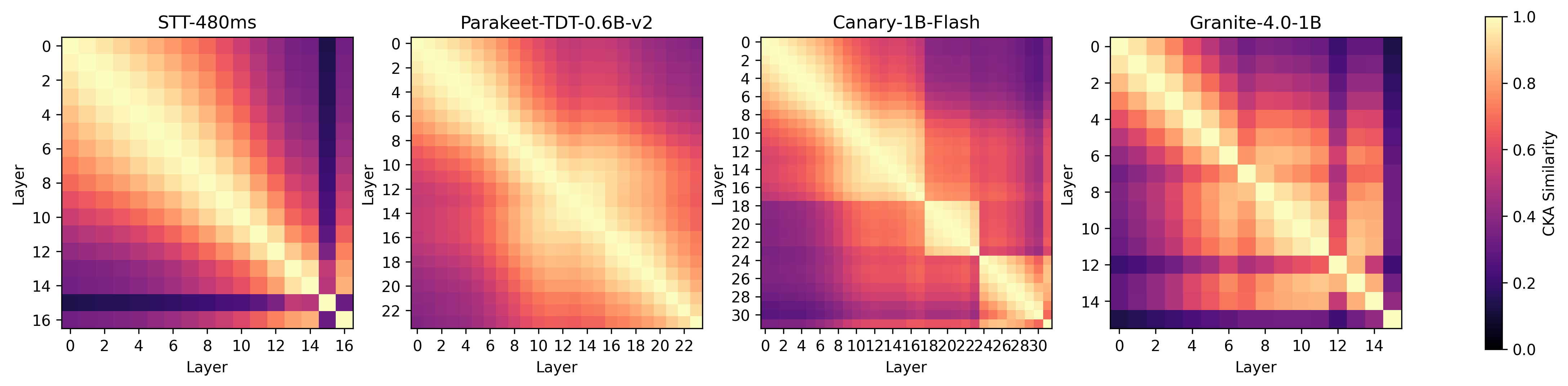} 
  \caption{Layer-wise CKA similarity for STT-480ms, Parakeet-TDT-0.6B-v2, Canary-1B-Flash, and Granite-4.0-1B-Speech encoder models.}
  \label{fig:cka}

  \label{fig:cka_angular}
\end{figure}

As shown in Figure \ref{fig:cka}, all four architectures in speech encoders exhibit strong block-diagonal structure in their CKA matrices, with layers forming groups of high similarity. A similar analysis of angular updates across successive layers (Appendix \ref{appendix:angular_dynamics}) reveals a consistent pattern, indicating that deep ASR encoders comprise groups of layers performing closely related transformations, rather than a sequence of entirely distinct computations. These findings suggest that the benefits of depth can be largely preserved by \emph{reusing a limited number of expressive blocks recurrently}, provided that the recurrence structure aligns with the model’s underlying  representational dynamics. To this end, we formalize this observation through the Block-Recurrent Encoder Hypothesis:
\paragraph{Definition 1 (Block-Recurrent Encoder Hypothesis)}
Let $f_L : \mathcal{X} \rightarrow \mathcal{H}$ denote a high-capacity ASR encoder composed of $L$ distinct layers, mapping an input acoustic feature sequence $x \in \mathcal{X}$ to intermediate representations. We say that $f_L$ satisfies the \emph{Block-Recurrent Hypothesis} if there exists $K \ll L$ encoder blocks $\{B_1,\dots,B_K\}$ and corresponding positive integers $\{n_1,\dots,n_K\}$ with $\textstyle\sum_{k=1}^{K} n_k = L$, such that \[\mathbb{E}_{x \sim \mathcal{D}}\left[\left\|f_L(x)-\left(B_K^{n_K} \circ \cdots \circ B_1^{n_1}\right)(x)\right\|_2\right]\leq \varepsilon,\]
where $B_k^{n_k}$ denotes $n_k$ recurrence of a single shared block $B_k$, $\forall k \in \{1, \dots, K \}$. $\mathcal{D}$ denotes the probability distribution over natural speech utterances, and $\varepsilon$ is a small positive value.

The hypothesis states that the computation performed by a deep ASR encoder can be closely approximated by a recurrent composition of a small number of shared blocks. Each block $B_k$ follows the standard Conformer design, consisting of feed-forward layers, multi-head self-attention, and convolutional modules, and is applied recurrently $n_k$ times. We empirically validate this hypothesis on multiple modern ASR encoders and construct compact models that satisfy it through representation-guided grouping and knowledge distillation \cite{kd}\cite{distillationdynamics}.

\section{Methodology}
In this section, we present the proposed block-recurrent ASR framework for accurate and efficient edge deployment. We first propose the construction of a block-recurrent model based on representation-guided layer grouping. We then introduce model initialization and training objective, including task-specific supervision, as well as encoder and decoder level knowledge distillation. Next, we show how varying the number of shared blocks yields a scalable family of edge-optimized models, allowing a single large ASR model to be distilled into multiple deployment-specific variants with different latency and memory budgets. We evaluate on-device latency and energy efficiency across diverse hardware platforms. Finally, we propose an on-device personalization extension based on MoE-LoRA, which adapts the compact block-recurrent backbone into a privacy-preserving user-specific model. Figure \ref{fig:block_recurrent_asr} provides an overview of the proposed workflow.

\subsection{Block-Recurrent ASR Encoder}

\paragraph{Definition 2 (Representation-guided Grouping)}
Given an ASR encoder of depth $L$, let $\mathbf{S} \in \mathbb{R}^{L \times L}$ denote a layer-wise representational similarity matrix, where $\mathbf{S}_{ij}$ is computed using linear CKA between activations of layers $i$ and $j$. A \emph{representation-guided grouping} is a partition of the encoder layers into $K$ ordered groups $\mathcal{G} = \{G_1, \dots, G_K\}$ ($K \ll L$), where each group $G_k$ corresponds to a contiguous segment layers, and is associated with a shared encoder block $B_k$. The grouping induces recurrence counts $\{n_1,\dots,n_K\}$ where $n_k = |G_k|$ and $\textstyle\sum_{k=1}^K n_k = L$.

In this work, we obtain $\mathcal{G}$ by maximizing intra‑group representational similarity in $\mathbf{S}$ using a dynamic programming procedure (Appendix \ref{appendix:algo2}). Each group $G_k$ therefore corresponds to a set of encoder layers that implement highly similar transformations, and can be well approximated by repeated application of a single shared block. The resulting repetition counts $\{n_k\}$ determine how many times block $B_k$ is applied in encoder, preserving the effective depth of the reference model while reducing the number of parameterized blocks from $O(L)$ to $O(K)$.

\subsection{Edge-ASR Construction and Training}
\label{sub:edge-asr}
\paragraph{Block initialization}
From the representation-guided grouping $\mathcal{G} = \{G_1,\dots,G_K\}$ obtained above, we construct a block-recurrent student encoder composed of $K$ shared blocks $\{B_1,\dots,B_K\}$. Each block $B_k$ is initialized by averaging the parameters of the corresponding teacher layers in group $G_k$. Specifically, let
$\theta_\ell$ denote the parameters of the $\ell$-th encoder layer in teacher model.
The parameters of the student block $B_k$ are initialized as:
\begin{equation}
    \label{eq:init}
    \theta_{B_k}\;=\;\tfrac{1}{|G_k|}\textstyle\sum_{\ell \in G_k} \theta_\ell \quad \forall k \in \{1, \dots, K\}.
\end{equation}

This initialization provides a stable starting point by aggregating parameters from representational similar layers, and empirically improves training stability and recognition performance compared to random initialization or using a single block's weights.

\paragraph{Training Objective}
The block-recurrent student model is trained using a combination of task-specific supervision and knowledge distillation from the corresponding teacher model. The training loss is defined as:
\begin{equation}
    \label{eq:loss} \mathcal{L}(\theta)=\mathcal{L}_{\text{task}}(\theta)+\lambda_{\text{enc}} \, \mathcal{L}_{\text{KD}}^{\text{enc}}(\theta)+\lambda_{\text{dec}} \, \mathcal{L}_{\text{KD}}^{\text{dec}}(\theta),
\end{equation}
where $\mathcal{L}_{\text{task}}$ denotes the task-specific ASR loss (e.g., CTC, RNN-T or TDT). $\mathcal{L}_{\text{KD}}^{\text{enc}}$ and $\mathcal{L}_{\text{KD}}^{\text{dec}}$ denote encoder level feature distillation loss and decoder level KL divergence loss, respectively. $\lambda_{\text{enc}}$ and $\lambda_{\text{dec}}$ are hyperparameter to control the relative contribution of each loss.

Note that the encoder level distillation is aligned with the representation-guided grouping, and is applied only at group boundaries. Specifically, for each group $G_k$ indexed from $s_k$ to $e_k$, we penalize the discrepancy between the student representation after the $n_k$-th recurrence of block $B_k$ and the teacher representation at layer $e_k$. 
\begin{equation}
    \label{eq:kd_enc}
\mathcal{L}_{\text{KD}}^{\text{enc}}
=\textstyle\sum_{k=1}^{K}\mathbb{E}_{x}\left[\left\|f_{e_k}^{T}(x)-f_{k}^{S}(x)\right\|_2\right].
\end{equation}
where $f_{e_k}^{T}(x)$ denotes the teacher encoder output at layer $e_k$, and $f_{k}^{S}(x)$ denotes the student encoder output after applying block $B_k$ for $n_k$ recurrences. Algorithm \ref{alg:block_recurrent_asr} summarizes the overall model construction and training procedure.

\paragraph{Scaling of Edge-ASR Models}
A key advantage of the block-recurrent formulation is that it enables a scalable family of edge-optimized ASR models derived from a single large-scale teacher. By varying the number of shared blocks $K$, we obtain encoders with different capacity levels: larger $K$ allocates more distinct blocks and improves accuracy, while smaller $K$ enforces stronger parameter sharing and yields compact models suitable for resource-constrained edge deployment. This allows a single large model to be distilled into multiple deployment-specific variants with different latency and memory budgets, while preserving the effective depth of the reference encoder. Furthermore, the block-recurrent design inherently benefits on-device execution: even under similar FLOP budgets, repeated weight reuse improves memory cache locality, leading to higher practical efficiency and lower end-to-end latency. We profile on-device latency and energy consumption across diverse hardware settings in Table \ref{tab:latency_grouping}, providing practitioners with guidance for selecting an appropriate block count $K$ that balances deployment constraints, recognition accuracy, and memory usage.

\subsection{On-Device Personalized ASR via MoE-LoRA}
\label{subsec:moe-lora} 
Under extreme compression regimes, especially when $K$ is small, the scaled block-recurrent model may exhibit a residual accuracy gap relative to the teacher due to limited encoder capacity. Importantly, some of the remaining gap appears to be user-dependent, with errors concentrated on speaker-specific characteristics such as accent, speaking style, and vocal timbre (Figure \ref{fig:per_user_wer_comparison_stt_group8_rank32}). To address user-specific errors, we introduce a lightweight personalization extension based on \emph{Mixture‑of‑Experts Low‑Rank Adaptation (MoE‑LoRA)}, which adapts the compact block-recurrent backbone to individual speakers without additional inference overhead.


\textbf{MoE-LoRA Adaptation}
During training, LoRA adapters are attached to linear layers within each block $B_k$, forming a small set of LoRA experts $\{\Delta W_1,\dots,\Delta W_N\}_k$, each specializing in different acoustic characteristics. A lightweight router with MLP layers conditions on a user embedding to select the top-$k$ experts, yielding a personalized update:

\begin{equation}
  \label{eq:lora}
    \Delta W(\mathbf{u}) = \textstyle\sum_{i \in \mathcal{T}_k(\mathbf{u})}
\alpha_i(\mathbf{u}) \, \Delta W_i,
\end{equation}
where $\mathbf{u}$ denotes the user embedding, $\mathcal{T}_k(\mathbf{u})$ denotes the index set of the top‑$k$ experts selected by the routing network, and weights $\alpha_i(\mathbf{u})$ are normalized over $\mathcal{T}_k(\mathbf{u})$. We jointly train the router and LoRA weights while keeping the block-recurrent encoder backbone fixed, and apply a load-balancing regularization term to prevent expert collapse. 

\textbf{Model Deployment}
At inference time, personalization begins with a brief on-device enrollment phase, during which a small amount of user speech is processed by an off-the-shelf speaker embedding model (e.g., ECAPA-TDNN \cite{ecapa}, CamPP \cite{campp}) to produce a user embedding, capturing speaker-specific characteristics. Crucially, embedding extraction is performed entirely on device, largely preserving user privacy. In our framework, the block-recurrent backbone remains frozen throughout personalization, preserving its general ASR capability. Personalization is introduced through lightweight user-conditioned MoE-LoRA adapters that provide additive, user-specific corrections without modifying the backbone parameters. After enrollment, the resulting LoRA updates can be fused into the base encoder at each block, allowing personalized ASR to execute with the same computation graph and latency as the backbone model. As a result, personalization incurs negligible runtime overhead while enabling the compact model to match or surpass the accuracy of a general-purpose ASR for individual users.

\section{Experiments and Results}
\label{sec:exp}
In this section, we evaluate the proposed block-recurrent ASR framework across multiple large-scale models, student capacities, and evaluation benchmarks. We first describe the experimental setup and evaluation protocol. We then analyze the effectiveness of representation-guided layer grouping, followed by a comprehensive evaluation of block-recurrent models on the Open ASR Leaderboard. We then demonstrate the performance of personalized ASR enabled by the compact block-recurrent backbone through MoE-LoRA adaptation. Finally, on-device latency and energy consumption are profiled across diverse hardware settings. We note that absolute recognition performance may vary with training hyperparameter, optimization schedules, and available training resources. Accordingly, our goal in this evaluation is not to exhaustively optimize each configuration for peak edge-ASR performance, but to demonstrate the effectiveness of structured recurrence as an architectural design principle, as well as its role in enabling efficient on-device personalization. 

\subsection{Experimental Setup}
\label{subsec:exp_setup}


\textbf{Large-scale Reference Models} We evaluate the proposed block-recurrent framework using two representative SOTA ASR models as case studies: STT-480ms and Parakeet-TDT-0.6B-v2, as summarized in Table \ref{appendix:teacher_models} in Appendix. The two models differ in encoder depth, decoder types, model sizes, and are representative of both \emph{streaming} and \emph{non-streaming} ASR paradigms, allowing us to assess the generality of the proposed approach across different deployment settings. 

\textbf{Training Datasets} Block-recurrent models are trained on a mixture of publicly available speech recognition datasets covering diverse acoustic conditions, domains, and utterance lengths. Utterance durations range from $0.1$s to $30$s, enabling models to learn from both short commands and long‑form conversational speech. Table \ref{tab:training_datasets} in Appendix \ref{appendix:dataset} summarizes the detailed datasets used in our experiments.

\textbf{Evaluation Benchmarks and Metrics} All models are evaluated using word error rate (WER) on benchmarks reported through the Open ASR Leaderboard \cite{openasr}, following its standardized evaluation protocol, official scoring scripts, and identical hardware configurations (Appendix \ref{appendix:hardware}). This ensures transparent, reproducible, and directly comparable results against existing ASR systems. 

\textbf{Implementation Details} All compact models preserve the effective encoder depth of the corresponding teacher model through block recurrence, and retain identical attention and convolutional configurations. Decoder architectures and text tokenizers are kept unchanged. Additional implementation details and training hyperparameter are provided in the Appendix \ref{appendix:implementation}.

\subsection{Block-Recurrent ASR Performances}
\label{subsec:student_result}
\textbf{Representation-guided Grouping} We evaluate the effectiveness of the proposed representation-guided grouping by comparing it against random contiguous grouping in a controlled ablation study. Using the STT-480ms teacher model and the LibriSpeech \texttt{clean} training set, we train block-recurrent student models with block budgets $K \in \{2,3,4,5\}$ under two settings: (i) grouping derived from the CKA similarity matrix, and (ii) randomly generated contiguous groupings. For the random baseline, we repeat training five times with different random partitions to capture performance variance. 

\captionsetup{font=footnotesize}
\begin{wrapfigure}{r}{0.45\linewidth}
  \vspace{-10pt}
  \centering
  \includegraphics[width=0.8\linewidth]{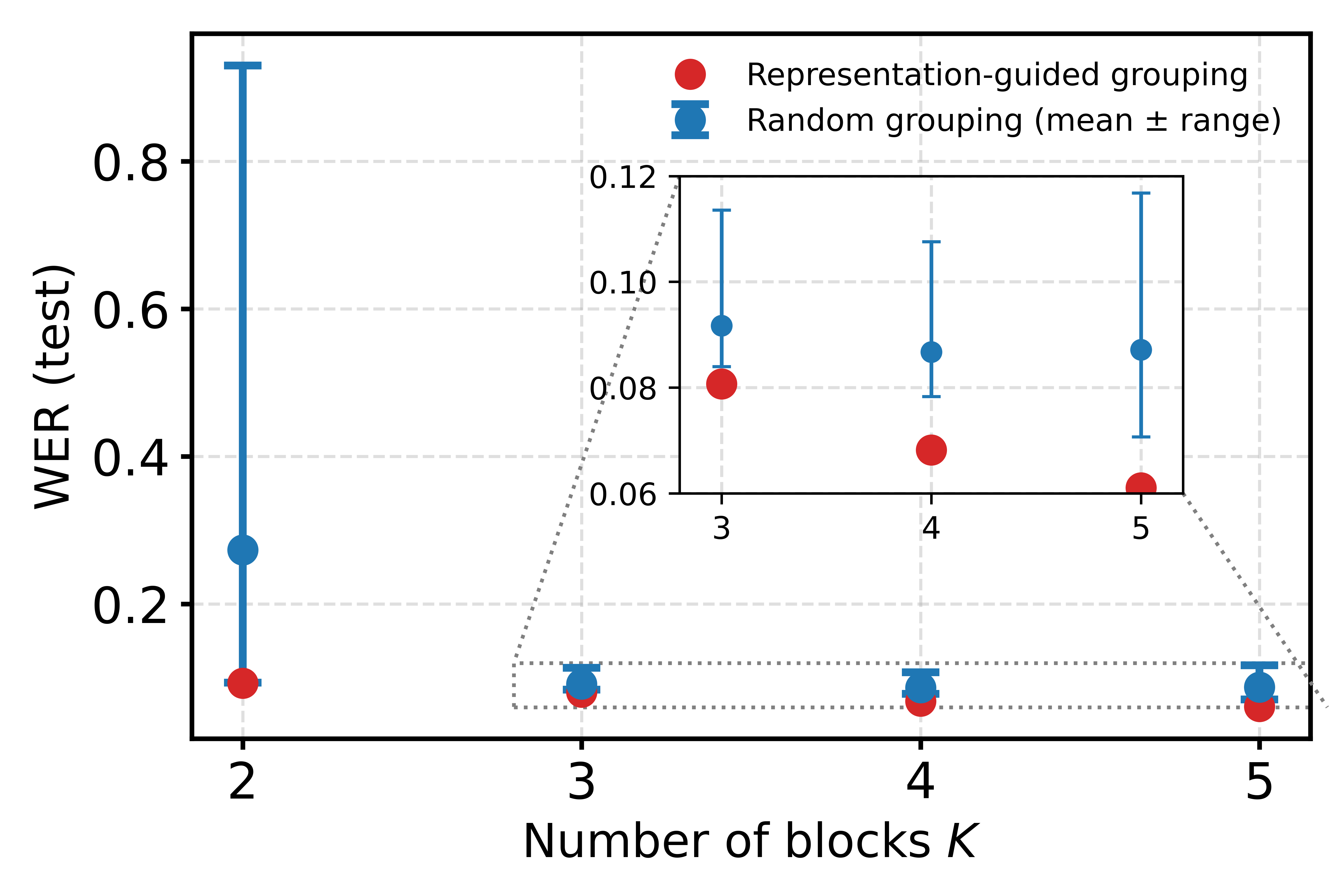}
  \vspace{-8pt}
  \caption{Representational grouping vs. random grouping (five runs with error bars).}
  \label{fig:grouping_vs_random}
\end{wrapfigure}

As shown in Figure~\ref{fig:grouping_vs_random}, representation-guided grouping consistently outperforms the mean performance of random grouping across all values of $K$, with particularly large gains under aggressive compression (e.g., $K=2$). These results demonstrate that representation-guided grouping is critical for effective block-recurrent modeling: reusing blocks without representational alignment leads to unstable and suboptimal performance, whereas CKA-based grouping yields consistently strong student models even under extreme parameter sharing.
\\[-12pt]

\textbf{Edge-ASR Performance on the Open ASR Leaderboard} We train block-recurrent models and evaluate performance on the Open ASR Leaderboard, following the standardized evaluation protocol \ref{subsec:exp_setup}. For each reference model, we report the teacher's performance alongside a family of block-recurrent student models distilled from the same teacher with varying block counts $K$, which control the degree of parameter sharing. All student models preserve the effective encoder depth through recurrence and are trained across all training datasets (Appendix Table \ref{tab:training_datasets}) using the same knowledge distillation and fine-tuning procedure. Note that the released teacher models were originally trained using substantially larger pipelines, including large-scale public/private corpora and multiple training stages. Several datasets in our 7-corpus mixture (e.g., LibriSpeech, AMI, VoxPopuli, and SPGISpeech) are already included in the teachers' original training data.

\begin{table*}[h]
\centering
\caption{Open ASR Leaderboard results for block-recurrent student models distilled from a single large-scale teacher. The teacher performance is reported in the first row in each section, followed by student models with varying numbers of shared blocks $K$. Encoder parameter counts are reported alongside the corresponding teacher encoder size to highlight the degree of parameter reduction.}
\label{tab:leaderboard_results}
\resizebox{\linewidth}{!}{%
\begin{tabular}{l cc cccccccc cc}
\toprule
& \multicolumn{2}{c}{\textbf{Configuration}} & \multicolumn{8}{c}{\textbf{Individual Datasets WER \%} $\downarrow$} & \multicolumn{2}{c}{\textbf{Overall WER \%} $\downarrow$} \\
\cmidrule(lr){2-3} \cmidrule(lr){4-11} \cmidrule(lr){12-13}
\textbf{Model} & \textbf{K} & \textbf{Params. (M)} & \textbf{AMI} & \textbf{Earn} & \textbf{Giga} & \textbf{LS-c} & \textbf{LS-o} & \textbf{SPGI} & \textbf{TED} & \textbf{VoxP} & \textbf{Avg.} & \boldmath$\Delta$ \\
\midrule
STT-480ms & \textendash & 108.9 & 26.23 & 20.56 & 14.15 & 2.54 & 5.97 & 5.80 & 5.58 & 7.60 & 11.05 & \textendash \\
\midrule
\multirow{5}{*}{Edge-STT}
 & 2  & 14.2 & 22.43 & 18.76 & 18.42 & 5.98 & 14.29 & 6.36 & 8.83 & 12.10 & 13.40 & \textcolor{red!75!black}{+2.35} \\
 & 4  & 26.8 & 18.16 & 15.24 & 15.60 & 4.29 & 10.55 & 4.72 & 7.34 & 9.89  & 10.72 & \textcolor{green!55!black}{-0.33} \\
 & 6  & 39.5 & 16.72 & 13.96 & 14.34 & 3.65 & 9.10  & 4.19 & 6.44 & 8.79  & 9.65  & \textcolor{green!55!black}{-1.40} \\
 & 8  & 52.1 & 14.93 & 13.17 & 13.36 & 3.25 & 8.08  & 3.76 & 6.10 & 8.38  & 8.88  & \textcolor{green!55!black}{-2.17} \\
 & 10 & 64.7 & 14.28 & 12.83 & 13.03 & 3.10 & 7.70  & 3.67 & 5.84 & 8.14  & 8.57  & \textcolor{green!55!black}{-2.48} \\
\midrule
Parakeet-v2 & \textendash & 608.9 & 11.16 & 11.09 & 9.73 & 1.70 & 3.19 & 2.18 & 3.37 & 5.96 & 6.05 & \textendash \\
\midrule
\multirow{5}{*}{Edge-Parakeet}
 & 4  & 105 & 13.28 & 11.93 & 12.21 & 2.66 & 6.40 & 2.68 & 4.60 & 7.16 & 7.62 & \textcolor{red!75!black}{+1.57} \\
 & 6  & 155 & 12.60 & 11.52 & 11.96 & 2.57 & 5.96 & 2.56 & 4.47 & 7.08 & 7.34 & \textcolor{red!75!black}{+1.29} \\
 & 8  & 205 & 12.21 & 11.35 & 11.63 & 2.31 & 5.48 & 2.44 & 4.30 & 6.85 & 7.07 & \textcolor{red!75!black}{+1.02} \\
 & 10 & 256 & 12.11 & 11.41 & 11.52 & 2.38 & 5.44 & 2.41 & 4.24 & 6.74 & 7.03 & \textcolor{red!75!black}{+0.98} \\
 & 12 & 306 & 11.81 & 11.31 & 11.43 & 2.31 & 5.30 & 2.36 & 4.17 & 6.75 & 6.93 & \textcolor{red!75!black}{+0.88} \\
\bottomrule
\end{tabular}}
\vspace{2pt}
{\scriptsize Earn=Earnings22, Giga=GigaSpeech, LS-c/o=LibriSpeech Clean/Other, SPGI=SPGISpeech, TED=TEDLIUM, VoxP.=VoxPopuli.}
\end{table*}

As shown in Table~\ref{tab:leaderboard_results}, increasing the number of shared blocks $K$ consistently improves recognition accuracy, reflecting the increased representational capacity of the block-recurrent encoder. This scaling allows practitioners to select an appropriate block count $K$ that meets specific on-device deployment requirements, balancing recognition accuracy against memory constraints. Notably, for STT-480ms model, a block-recurrent student with only \textbf{two shared Conformer blocks} recovers \textbf{97.4\%} of the teacher’s overall recognition accuracy, despite an order-of-magnitude reduction in encoder parameters. With four shared blocks, the model already matches the teacher’s average WER. This trend generalizes across architectures. Despite substantial differences in encoder depth and model sizes between STT-480ms and Parakeet-TDT-0.6B-v2, block-recurrent student models exhibit similar compression-accuracy trade-offs in both cases. These results suggest a general property: a substantial fraction of the performance gains typically attributed to deep ASR encoders can be preserved through structured recurrence over a small set of shared blocks.

\subsection{Personalized ASR Performance}
We evaluate personalized ASR performance across different LoRA ranks under $16$ experts built on the block-recurrent backbone, as described in Section~\ref{subsec:moe-lora}. Experiments are conducted on VoxCeleb2, a speaker-labeled dataset for which transcriptions are generated through a rigorous annotation process (Appendix~\ref{subsubsec:vox2_dataset}). For each speaker, the user embedding selects the top-$5$ LoRA experts. 
\vspace{-10pt}

\captionsetup[table]{skip=0pt}
\begin{table}[h]
\caption{Performance comparison of teacher models (first row), block-recurrent, and personalized models with varying MoE-LoRA ranks on the VoxCeleb2, averaged over $106$ test speakers. Zero-shot denotes ASR without MoE-LoRA personalization.}
\centering
\begin{subtable}{0.45\linewidth}
\footnotesize
\centering
\caption{Edge-STT (Test WER \% $\downarrow$)}
\renewcommand{\arraystretch}{0.5}
\begin{tabular}{ccccc}
\toprule
\multirow{2}{*}{$K$} & \multirow{2}{*}{Zero-shot} & \multicolumn{3}{c}{MoE-LoRA (Rank)} \\
\cmidrule(lr){3-5}
& & 16 & 32 & 64 \\
\midrule
\textendash & 11.80 & \textendash & \textendash & \textendash \\
\midrule
4 & 15.60 & 11.72 & 11.37 & 11.17 \\
6 & 13.51 & 9.06  & 8.91  & 8.88  \\
8 & 12.19 & 8.18  & 8.03  & 8.04  \\
\bottomrule
\end{tabular}
\end{subtable}
\hfill
\begin{subtable}{0.45\linewidth}
\footnotesize
\centering
\caption{Edge-Parakeet (Test WER \% $\downarrow$)}
\renewcommand{\arraystretch}{0.5}
\begin{tabular}{ccccc}
\toprule
\multirow{2}{*}{$K$} & \multirow{2}{*}{Zero-shot} & \multicolumn{3}{c}{MoE-LoRA (Rank)} \\
\cmidrule(lr){3-5}
&  & 16 & 32 & 64 \\
\midrule
-- & 4.35 & -- & -- & -- \\
\midrule
4 & 8.92 & 6.19 & 6.07 & 6.31 \\
6 & 8.74 & 5.66 & 5.53 & 5.51 \\
8 & 8.12 & 5.24 & 5.20 & 5.10 \\
\bottomrule
\end{tabular}
\end{subtable}
\label{tab:moe-lora-results}
\end{table}
\vspace{-5pt}

Table~\ref{tab:moe-lora-results} compares performance across the general-purpose teacher model, block-recurrent models, and their personalized counterpart augmented with MoE-LoRA adapters. Overall, lightweight MoE-LoRA adaptation substantially improves user-specific recognition accuracy over the non-persoanlized block-recurrent baseline across all values of $K$. For a fixed number of recurrent blocks, increasing the LoRA rank generally yields better performance, with the largest gains obtained when scaling the rank from $16$ to $32$ in most settings. Notably, for the Edge-STT model, LoRA adaptation with rank 16 effectively closes the accuracy gap between the on-device personalized model and the general-purpose system. For Edge-Parakeet, the best result is achieved with $K=8$ and rank $64$, reaching $5.10\%$ WER and reducing the gap to the reference model to $0.75\%$.

\vspace{-8pt}
\begin{figure*}[h]
    \centering
    \includegraphics[width=0.95\textwidth]{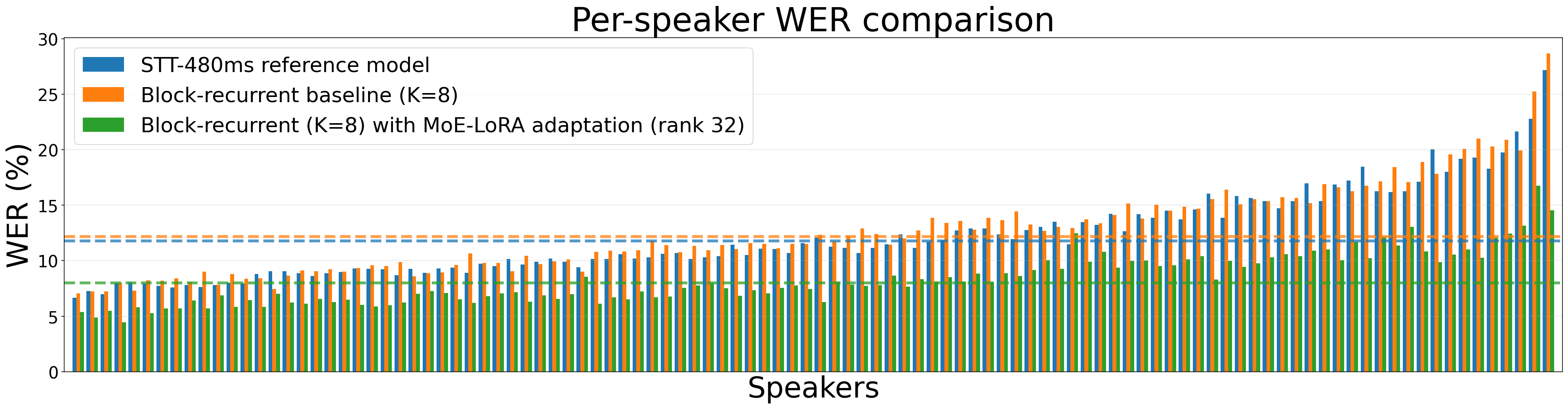}
    \caption{
    Per-speaker WER comparison for 106 test speakers, among the STT-480ms reference model, the block-recurrent baseline ($K=8$), and the personalized MoE-LoRA model (rank 32). Each grouped bar represent an individual speaker. Dashed horizontal lines, shown in the corresponding colors, indicate the overall pooled WER for each model.}
\label{fig:per_user_wer_comparison_stt_group8_rank32}
\end{figure*}

Figure~\ref{fig:per_user_wer_comparison_stt_group8_rank32} shows per-speaker zero-shot performance for the STT-480ms reference model, the block-recurrent baseline ($K=8$), and the corresponding MoE-LoRA model across $106$ test speakers. The reference model exhibits large variation (WER $5\%-28\%$), indicating that ASR performance may be influenced by user-specific factors such as accent and speaking style. In contrast, the personalized MoE-LoRA model substantially reduces WER for every speaker, and achieves a much lower overall pooled WER. These results demonstrate that lightweight, speaker-specific adaptation is effective at the individual-user level, making the edge-scale ASR model more robust to speaker-level variability.

\subsection{Inference Efficiency of Block-Recurrent ASR Models}

To empirically validate the inference efficiency due to improved cache locality introduced by the block-recurrent design. In Table~\ref{tab:latency_grouping}, we report the latency for the STT‑480ms encoder ($L{=}17$, $T{=}77$ frames) under block-recurrent configurations with varying numbers of blocks $K$. In all configurations, the effective encoder depth is preserved at $17$ layers, matching the reference model through recurrent application of shared blocks. Across both GPU and CPU platforms, block recurrence reduces effective memory traffic through weight reuse, resulting in measurable improvements in end‑to‑end latency despite comparable FLOP counts. On GPU, block-recurrent models achieve up to a $1.21{\times}$ latency reduction in memory-bound regimes (scalar kernel) and a stable $1.06{\times}$ improvement in compute-bound regimes (dot kernel). On CPU, block recurrence yields a more modest $1.03\text{--}1.07{\times}$ speedup under different cache configurations across both server-class and consumer-grade processors. Full profiling details and energy measurements are provided in Appendix~\ref{appendix:latency}. It is noted that existing accelerators and runtime systems are not optimized for recurrent weight reuse. Consequently, the observed gains under standard implementations likely underestimate the potential benefits of block recurrence. As shown in Table~\ref{tab:latency_grouping}, cache-aware execution and larger cache capacities already yield more pronounced latency improvements, suggesting that hardware and software stacks designed to exploit weight reuse could further improve runtime efficiency.

\begin{table*}[h]
\centering
\footnotesize
\caption{Inference latency (ms) and speedup for block-recurrent STT model under various block counts $K$. GPU inference uses custom Triton kernels in two variants: a scalar kernel that computes per-token outer products and operates primarily in a memory-bound regime, and a dot kernel based on \texttt{tl.dot}, which uses tensor-core-accelerated tiled matmul. 
}
\label{tab:latency_grouping}
\resizebox{\linewidth}{!}{
\begin{tabular}{cc cc ccc}
\toprule
& & \multicolumn{2}{c}{\textbf{GPU} (Blackwell RTX Pro 6000)}
  & \multicolumn{3}{c}{\textbf{CPU} (fp32)} \\
\cmidrule(lr){3-4} \cmidrule(lr){5-7}
\textbf{Model} & $K$
  & Scalar (fp16) & Dot (fp16)
  & Threadripper 7995WX & Apple M4 Pro & \\
\midrule
STT-480ms & \textendash
 & $6.22\pm.01\ (1.00\times)$
 & $3.88\ (1.00\times)$
 & $55.7\pm.43\ (1.00\times)$
 & $43.44\pm.34\ (1.00\times)$\\
\midrule
\multirow{5}{*}{Edge-STT}
 & 1
 & $5.13\pm.02\ (\mathbf{1.21\times})$
 & $3.67\ (\mathbf{1.06\times})$
 & $52.0\pm.44\ (\mathbf{1.07\times})$
 & $42.26\pm.20\ (\mathbf{1.03\times})$\\
 & 2
 & $5.21\pm.02\ (1.19\times)$
 & $3.68\ (\mathbf{1.06\times})$
 & $52.0\pm.39\ (\mathbf{1.07\times})$
 & $42.35\pm.25\ (\mathbf{1.03\times})$\\
 & 4
 & $5.27\pm.01\ (1.18\times)$
 & $3.70\ (1.05\times)$
 & $52.8\pm.35\ (1.05\times)$
 & $42.52\pm.15\ (1.02\times)$\\
 & 6
 & $5.59\pm.01\ (1.11\times)$
 & $3.73\ (1.04\times)$
 & $52.9\pm.33\ (1.05\times)$
 & $42.75\pm.33\ (1.02\times)$\\
 & 8
 & $5.93\pm.02\ (1.05\times)$
 & $3.76\ (1.03\times)$
 & $54.1\pm.90\ (1.03\times)$
 & $42.81\pm.22\ (1.01\times)$ \\
\bottomrule
\end{tabular}}
\end{table*}
\vspace{-5pt}

\section{Limitations and Future Work}
\label{future_work}
While the proposed block-recurrent ASR framework demonstrates strong empirical performance, it also has limitations that point to promising directions for future work.

\textbf{Limitations} Our representation-guided grouping relies on CKA computed from layer activations. While CKA is widely used for measuring representational similarity, it involves normalization steps that may be sensitive to outliers or distributional shifts in activation statistics. In extreme cases, this sensitivity could affect the robustness of the resulting layer groupings. Exploring more robust variants of CKA is a promising direction for improving grouping stability under challenging conditions. 

\textbf{Future work} In this work, block-recurrent models use a fixed number of iterations derived from grouping to preserve the effective depth of the teacher model while maintaining comparable FLOPs. A natural extension is to incorporate \emph{adaptive halting mechanisms} that allow the model to dynamically determine the required number of recurrent steps for each block on a per-utterance basis. Such mechanisms could further reduce inference computation in addition to parameter counts by allocating fewer iterations to acoustically simpler inputs while retaining sufficient iterations for more challenging conditions. We leave the integration of adaptive halting and its impact on accuracy-efficiency trade-offs to future work. In addition, while this work focuses on architectural parameter sharing, block-recurrent structures are orthogonal to model quantization techniques. Combining block recurrence with low-bit quantization could further reduce memory footprint and improve inference efficiency for on-device deployment. Beyond ASR, the observed representational dynamics may extend to other sequence modeling tasks based on deep Transformer-style encoders or decoders, and exploring whether block-recurrent architectures can similarly preserve performance while reducing model size in these domains remains an important direction for future research.


\section{Conclusion}
\label{sec:conclusion}
This work revisits the role of depth in modern ASR encoders and demonstrates that high-accuracy on-device ASR can be achieved without large parameterization. Through representational analysis of state-of-the-art ASR models, we observe that many layers learn functionally similar transformations, and can therefore be approximated through iterative refinement. Building on this insight, we introduced a block-recurrent ASR encoder that replaces parameterized depth with recurrent application of a small number of shared blocks. Using representation-guided grouping and targeted knowledge distillation, block-recurrent models preserve the effective depth and recognition accuracy of large ASR systems while reducing model size by an order of magnitude. Extensive evaluations across multiple large-scale teacher models and benchmarks demonstrate that compact block-recurrent students achieve competitive word error rates in both streaming and non-streaming settings. Beyond compression, we enable practical on-device personalization on the block-recurrent backbone through lightweight MoE-LoRA adaptation, yielding user-specific accuracy gains without additional inference cost. Overall, our results suggest that structured recurrence provides a principled path towards ultra-compressed, energy-efficient, high-quality, and personalized speech recognition on the edge.

\newpage

\bibliographystyle{unsrt}
\bibliography{neurips_2026}

\newpage

\appendix

\section{Implementation Details}
\label{appendix:implementation}

This appendix provides detailed information on the implementation and training setup used in our experiments.

\subsection{Training Hardware Details}
\label{appendix:hardware}
Throughout our training work, we use the environment setup with NVIDIA A100 GPU (40/80GB), PyTorch 2.9.1 framework, CUDA version 12.4, and Ubuntu 22.04 LTS. 

\subsection{Large-scale Model Details}
\label{appendix:teacher_models}
Table \ref{tab:teacher_model_breakdown} provides a detailed encoder-decoder breakdown of the four models analyzed in Figure \ref{fig:cka}. Our CKA similarity analysis is restricted to the acoustic encoder, as it is the primary component targeted for on-device ASR deployment and personalization. Our training and experimental evaluation consider two representative ASR models \ref{tab:teacher_model_hyperparams}: NVIDIA STT-480ms \cite{stt} and Parakeet-TDT-0.6B-v2 \cite{parakeet}. NVIDIA STT‑480ms is a streaming FastConformer‑CTC/RNN-T ASR system optimized for sub-second latency. Parakeet-TDT-0.6B-v2 is a large-scale offline Conformer-TDT model trained using the NeMo toolkit. The model details are listed in the following.

\begin{table}[h]
\centering
\caption{ASR models: Enc/Dec breakdown}
\resizebox{1.0\linewidth}{!}{
\begin{tabular}{c c c c c c}
\hline
\textbf{Model} & \textbf{Encoder} & \textbf{Encoder Params.} & \textbf{Decoder} & \textbf{Decoder Params.} & \textbf{Enc/Dec Param Ratio}\\
\hline
STT-480ms & 17 FastConformer blocks & 108.9M & RNNT/CTC head & 5.9M & 18.5x \\
Parakeet-v2 & 24 Conformer blocks & 608.9M & TDT head & 8.9M & 68.4x \\
Canary-1B-Flash & 32 FastConformer blocks & 883.6M & 4 Transformer blocks & 72.6M & 12.1x \\
Granite-4.0-1B-Speech & 16 Conformer blocks & 440.2M & LLM decoder & 1.8B & 0.24x \\
\hline
\end{tabular}}
\label{tab:teacher_model_breakdown}
\end{table}

\begin{table}[h]
\centering
\caption{Large-scale teacher models}
\resizebox{0.9\linewidth}{!}{
\begin{tabular}{c c c}
\hline
\textbf{Model} & \textbf{STT-480ms} & \textbf{Parakeet-TDT-0.6B-v2} \\
\hline
encoder layers & 17 & 24 \\
hidden dim. & 512 & 1024 \\
encoder type & FastConformer with casual chunked attention & Conformer \\
att\_context\_size & [70, 6]   & \textendash \\
decoder type & hybrid CTC/RNN-T & TDT \\
tokenizer & BPE & BPE \\
vocabulary size & 1024 & 1024 \\
encoder params. & $108.9$M & $608.9$M \\
model params. & $114.8$M & $617.8$M \\
streaming & Yes & No \\
\hline
\end{tabular}}
\label{tab:teacher_model_hyperparams}
\end{table}

\subsection{Block-Recurrent Model Details}
\label{appendix:teacher_models}
Table \ref{tab:stt_student} reports the parameter counts of our block-recurrent students and the compression rate expressed as a percentage of parameter reduction relative to the corresponding teacher model. 
\begin{table}[h]
\centering
\caption{Scaling block-recurrent student models}
\begin{subtable}{\linewidth}
\centering
\begin{tabular}{ c c c }
\hline
\multicolumn{3}{c}{\textbf{STT-480ms}}\\
\hline
Block count K & Encoder params. & Parameter reduction rate  \\
\hline
2 & 14.2 & 87.0\% \\
4 & 26.8 & 75.4\% \\ 
6 & 39.5 & 63.7\% \\
8 & 52.1 & 52.2\% \\
10 & 64.7 & 40.6\% \\
\end{tabular}
\end{subtable}
\vspace{0.5em}
\begin{subtable}{\linewidth}
\centering
\begin{tabular}{ c c c }
\hline
\multicolumn{3}{c}{\textbf{Parakeet-TDT-0.6B-v2}}\\
\hline
Block count K & Encoder params. & Parameter reduction rate \\
\hline
4 & 105 & 82.9\%\\ 
6 & 155 & 74.9\%\\ 
8 & 205 & 66.8\%\\ 
10 & 256 & 58.5\%\\ 
12 & 306 & 50.4\%\\ 
\hline
\end{tabular}
\end{subtable}
\label{tab:stt_student}
\end{table}

\subsection{Representation Grouping Details}
Table \ref{tab:grouping} summarizes the representation-guided layer groupings produced by our algorithm \ref{alg:dp} for each model. 
\begin{table}[H]
\centering
\caption{Representational grouping details}
\resizebox{\linewidth}{!}{
\begin{tabular}{c c l}
\hline
\textbf{Model} & \textbf{Block count K} & \textbf{Grouping (start layer index, end layer index)} \\
\hline
 \multirow{5}{*}{STT-480ms} & 2 & [(1, 16), (17, 17)] \\
 & 4 & [(1, 13), (14, 15), (16, 16), (17, 17)] \\
 & 6 & [(1, 11), (12, 12), (13, 13), (14, 15), (16, 16), (17, 17)] \\
 & 8 & [(1, 8), (9, 11), (12, 12), (13, 13), (14, 14), (15, 15), (16, 16), (17, 17)] \\
 & 10 & [(1, 4), (5, 7), (8, 9), (10, 11), (12, 12), (13, 13), (14, 14), (15, 15), (16, 16), (17, 17)] \\
\hline
\multirow{5}{*}{Parakeet-TDT-0.6B-v2} & 4 & [(1, 5), (6, 8), (9, 23), (24, 24)] \\
& 6 & [(1, 5), (6, 8), (9, 21), (22, 22), (23, 23), (24, 24)] \\
& 8 & [(1, 3), (4, 6), (7, 8), (9, 9), (10, 21), (22, 22), (23, 23), (24, 24)] \\
& 10 & [(1, 3), (4, 6), (7, 8), (9, 9), (10, 19), (20, 20), (21, 21), (22, 22), (23, 23), (24, 24)] \\
& 12 & [(1, 3), (4, 6), (7, 8), (9, 9), (10, 10), (11, 17), (18, 19), (20, 20), (21, 21), (22, 22), (23, 23), (24, 24)] \\
\hline
\end{tabular}}
\label{tab:grouping}
\end{table}

\section{Training Dataset and Hyperparameter}
\label{appendix:dataset}

\subsection{OpenASR Datasets}
Table \ref{tab:training_datasets} summarizes the mixture of publicly available speech recognition datasets used for distilling the block-recurrent student model. Specifically, we use training splits of LibriSpeech \cite{dataset_lib}, TEDLIUM Release-3 \cite{dataset_ted}, VoxPopuli \cite{dataset_vox}, AMI \cite{dataset_ami}, Earnings22 \cite{dataset_earnings}, GigaSpeech \cite{dataset_giga}, and SPGISpeech \cite{dataset_spgi}. The test sets strictly follow the Open ASR Leaderboard protocol. Unlike the STT-480ms model, which is case-insensitive at the tokenizer level, the Parakeet-TDT-0.6B-v2 model is case-sensitive. This creates a challenge for knowledge distillation as ground-truth transcripts follow varying casing conventions, while Parakeet teacher predictions contain mixed case, leading to potential token-ID mismatches despite textual equivalence. To mitigate this issue, for Edge-Parakeet, we transcribe the training data using the Parakeet teacher model and replace the original ground-truth transcripts with the teacher transcriptions for samples whose WER is $\leq 0.2$. This filtering retains approximately $71.3\%\text{--}99.7\%$ of the original dataset.

\begin{table}[h]
\centering
\caption{Training datasets used for distilling block-recurrent models.}
\label{tab:training_datasets}
\resizebox{1.0\linewidth}{!}{
\begin{tabular}{lcccc}
\toprule
\textbf{Dataset} & \textbf{Domain} & \textbf{Utterance Length} & \textbf{Samples for STT (train/val.)} & \textbf{Samples for Parakeet (train/val.)}\\
\midrule
LibriSpeech        & Read speech          & 0.8-30 s & 281242 / 5551  & 278978 / 5417\\
Tedlium v3         & Talks / lectures     & 0.5-15 s & 265274 / 500  & 249141 / 486\\
VoxPopuli          & Parliamentary speech & 0.5-30 s & 181050 / 1742  & 173958 / 1647\\
AMI                & Meetings             & 0.5-30 s & 73141 / 9196  & 56627 / 7090\\
Earnings22         & Financial calls      & 1-30 s & 47564 / 2456  & 33906 / 1849\\
GigaSpeech         & Read + spontaneous   & 0.5–20 s & 910140 / 5688  & 856002 / 4656\\
SpgiSpeech         & Spontaneous speech   & 5–15 s & 385361 / 39304  & 384203 / 39165\\
\bottomrule
\end{tabular}}
\end{table}

\subsection{VoxCeleb2 Dataset}
\label{subsubsec:vox2_dataset}
Table \ref{tab:personalized_datasets} summarizes the dataset splits for VoxCeleb2. To obtain reliable ground-truth transcriptions, we applied a rigorous curation pipeline. First, each audio sample was transcribed using two state-of-the-art OpenASR leaderboard models: Canary-Qwen-2.5B and Parakeet-TDT-0.6B-v2. The resulting transcripts were normalized, and the word error rate (WER) etween the two outputs was computed. Based on manual inspection, we establish a WER threshold of $0.15$ and discard any utterances where the WER between the two models exceeded the threshold. To ensure sufficient training data per speaker, we then removed speakers with less than $8$ minutes of retained audio after filtering. Finally, We adopted the Canary-Qwen-2.5B transcripts as the final ground truth, given its superior capabilities. This strict selection process resulted in a $47.32\%$ speaker retention rate, yielding a refined dataset of $2892$ speakers from the original $6112$. Since, the speakers in the original Voxceleb2 Dataset is split into a dev and a test set, we divide the dev set into train and validation with an approximate 90/10 user split. Final evaluation is performed by averaging results over $106$ test speakers.



\begin{table}[h]
\centering
\caption{Dataset split of Voxceleb2 for personalized ASR.}
\begin{tabular}{ c c c }
\hline
\textbf{Split} & \textbf{\# of Speakers} & \textbf{Total duration (min)} \\
\hline
train & 2507 & 70452 \\ 
dev & 279 & 8037 \\ 
test & 106 & 3986 \\
\hline
\end{tabular}
\label{tab:personalized_datasets}
\end{table}

\newpage 

\subsection{Training Hyperparameter}
Table \ref{tab:training_hyperparams} shows the major hyperparameter used across the experiments. Pre-training the block-recurrent student models takes approximately 7-10 days on two NVIDIA A100 GPUs for both model variants. Subsequent fine-tuning for 20 epochs can be completed within 3 days under the same hardware configuration. All models are trained under consistent settings to ensure fair comparison.   

\begin{table}[H]
\centering
\caption{Training hyperparameter}
\begin{tabular}{c c}
\hline
\textbf{Hyperparameter} & \textbf{Value} \\
\hline
batch size & 8 \\ 
learning rate (LR) & 5e-4/5e-5 \\ 
LR scheduler & constant \\
optimizer & AdamW \\
$\beta_1$ & 0.9 \\
$\beta_2$ & 0.98 \\
weight decay & 1e-3/1e-4 \\
epochs & 50 (pre-training), 20 (fine-tuning) \\
mixed precision & bf16 \\ 
\hline
\end{tabular}
\label{tab:training_hyperparams}
\end{table}

Table \ref{tab:training_hyperparams_moe_lora} illustrates the hyperparameter used for each MoE-LoRA training. Each training takes around 3-4 days on two NVIDIA A100 GPUs for the STT-480ms edge family, and four NVIDIA A100 GPUs for the Parakeet-v2 edge family. 

\begin{table}[H]
\centering
\caption{Training hyperparameter for MoE-LoRA}
\begin{tabular}{c c}
\hline
\textbf{Hyperparameter} & \textbf{Value} \\
\hline
batch size & 8 \\ 
learning rate (LR) & 5e-4 \\ 
LR scheduler & constant \\
optimizer & AdamW \\
$\beta_1$ & 0.9 \\
$\beta_2$ & 0.98 \\
weight decay & 1e-3 \\
epochs & 30 \\
mixed precision & bf16 \\ 
total number of experts & 16 \\
k & 5 \\
number of top experts selected & 5 \\
LoRA rank & 16/32/64 \\
Router mlp depth & 2\\
Router mlp hidden dimension & 256\\

\hline
\end{tabular}
\label{tab:training_hyperparams_moe_lora}
\end{table}

\newpage 

\section{Algorithm Details}
\label{appendix:algorithm}

Algorithm \ref{alg:block_recurrent_asr} summarizes the procedure for converting a large-scale ASR model into its block-recurrent design, while Algorithm \ref{alg:dp} details the CKA-based representation layer grouping. The CKA protocol used to compute the similarity matrix is described in \ref{appendix:cka_protocol}.

\subsection{Block-Recurrent Edge-ASR Model}
\label{appendix:algo1}

\begin{algorithm}[h]
\caption{Block-Recurrent Model Distillation}
\label{alg:block_recurrent_asr}
\begin{algorithmic}

\State {\bfseries Input:} Teacher ASR model $f_T$,
calibration set $\mathcal{D}_C$,
training set $\mathcal{D}_T$,
block budget $K$, 
training budget $T_1$, $T_2$,
learning rate $\eta$
\State {\bfseries Output:} Block-recurrent student model $f_S$
\\

\State Compute similarity matrix $S$ with activations $f_T(\mathcal{D}_C)$
\State Derive layer groups $\{G_1,\dots,G_K\}$ given $S$, $K$ via Algorithm \ref{alg:dp}

\State Initialize student encoder $B_k \gets {\mathrm{Avg}}(G_k),\ \forall k$ via Eq.~(\ref{eq:init}) and freeze decoder
\\

\For{$T_1$ epochs}
\For{batch $x \sim \mathcal{D}_T$}
\State Compute task loss $\mathcal{L}_{task}(x)$, decoder loss $\mathcal{L}_{KD}^{dec}(x)$, encoder loss $\mathcal{L}_{KD}^{enc}(x)$ as Eq. \ref{eq:loss}, \ref{eq:kd_enc}
\State Update student $\theta_S \gets \theta_S - \eta \nabla_{\theta_S} \mathcal{L}$
\EndFor
\EndFor

Optionally fine-tune student encoder and decoder together for $T_2$ epochs.

\end{algorithmic}
\end{algorithm}

\subsection{Representation-Guided Grouping}
\label{appendix:algo2}

\begin{algorithm}[h]
\caption{CKA-based Representation Layer Grouping}
\label{alg:dp}
\begin{algorithmic}

\State {\bfseries Input:} Similarity matrix $S \in \mathbb{R}^{L \times L}$;
number of blocks $K$
\State {\bfseries Output:} Contiguous layer groups $\{G_1,\dots,G_K\}$
\\

\For{$1 \le i \le j \le L$}
  \State Compute similarity score $C_{ij} \gets
  \dfrac{
    \sum_{p=i}^{j}\sum_{q=i}^{j} S_{pq}
  }{(j-i+1)^2}$
\EndFor
\\ 

\State initialize $\mathrm{dp}[1,j] \gets C_{1j} , \forall j \in {1, \dots L}$
\State $\triangleright$ \textit{let $\mathrm{dp}[k,j]$ be the best score for partitioning $1,\dots,j$ into $k$ segments}
\\

\For{$k = 2$ \textbf{to} $K$}
  \For{$j = k$ \textbf{to} $L$}
    \State
    $\mathrm{dp}[k,j] \gets
    \max_{i \in \{k-1,\dots,j-1\}}
    ( \mathrm{dp}[k-1,i] + C_{i+1,j}
    $ )
    \State $\triangleright$ \textit{also store argmax $i$ for backtracking}
  \EndFor
\EndFor
\State Recover boundaries $\{G_1,\dots,G_K\}$ via backtracking

\end{algorithmic}
\end{algorithm}

\subsection{Linear CKA Protocol}
\label{appendix:cka_protocol}

For each calibration utterance, we extract hidden activations from every encoder layer, mean-center them along the temporal dimension, and compute pairwise linear CKA scores, producing an $L \times L$ similarity matrix. CKA between layer representations X and Y is computed as:

\begin{equation}
\text{S}(X, Y) = \frac{\|X^{\top}Y\|_F^2}{\|X^{\top}X\|_F \cdot \|Y^{\top}Y\|_F}
\end{equation}
where $\|\cdot\|_F$ denotes the Frobenius norm. To obtain a robust estimate of representational redundancy, we average these matrices over a calibration set of $150$ utterances, consisting of $50$ samples from the development splits of each of three acoustically and linguistically diverse corpora: LibriSpeech, TED-LIUM, and VoxPopuli. The resulting averaged heatmap is used to identify groups of layers whose representations remain strongly aligned across datasets and speaking conditions.

\section{Additional Experimental Results and Analysis}
\label{appendix:additional_analysis}

\subsection{Grouping Sensitivity Analysis}
\label{appendix:cka_sensitivity}
To assess the robustness of the resulting groupings to the choice of calibration data, we conducted an ablation study in which CKA matrices were computed using each of the three calibration corpora independently. For the STT-480ms model, the resulting CKA-based groupings were highly consistent across all three individual calibration datasets. For Parakeet-v2, only minor variations were observed in a small number of group boundaries, while the overall grouping structure remained unchanged. 

To quantify the stability, we compute pairwise Pearson correlations between the CKA similarity matrices derived from each calibration dataset. Specifically, an $L \times L$ CKA similarity matrix is computed independently for each dataset. For every dataset pair, the corresponding matrices are flattened into vectors, and their Pearson correlation coefficient is calculated:

\begin{table}[H]
\centering
\caption{Cross-dataset Pearson correlations of CKA similarity matrices}
\resizebox{0.6\linewidth}{!}{%
\begin{tabular}{c c c c}
\hline
\textbf{Model} & \textbf{LS vs TED} & \textbf{LS vs VoxP} & \textbf{TED vs VoxP}\\
\hline
STT-480ms & 0.9989	& 0.9977	& 0.9992 \\
Parakeet-v2 & 0.9969 & 0.9943   & 0.9945 \\
\hline
\end{tabular}}
\label{tab:pearson}
\end{table}
High correlations (>0.99) across all dataset pairs suggest that the CKA-guided grouping structure is highly stable across calibration data sources.

\subsection{Layer-wise Representational Dynamics}
\label{appendix:angular_dynamics}

\begin{figure}[h]
  \centering
\includegraphics[width=1.0\linewidth,height=0.15\textheight,keepaspectratio]{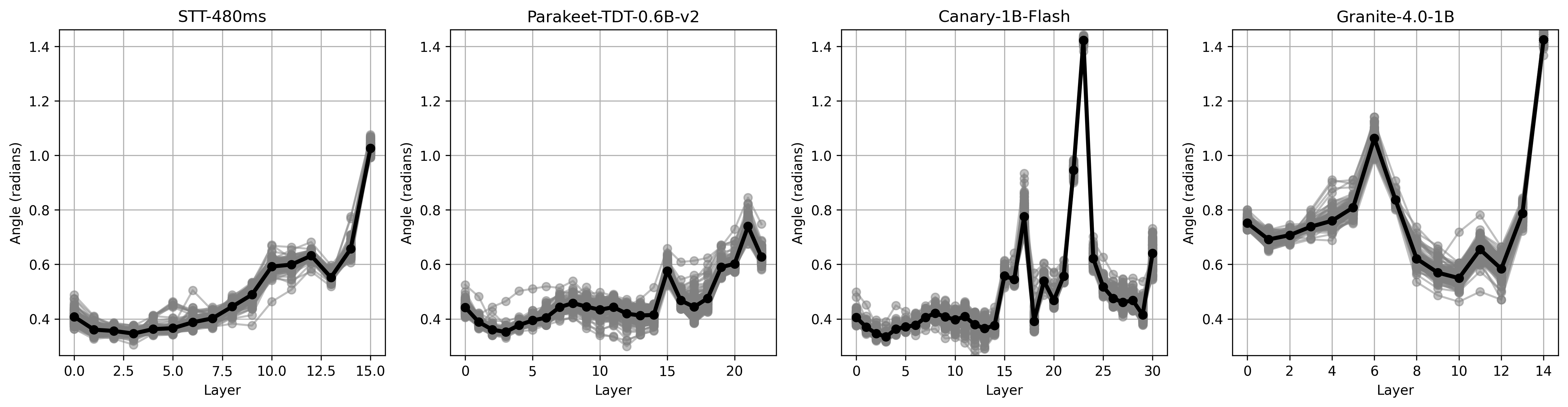} 
  \caption{Angular updates across successive layers for STT-480ms, Parakeet-TDT-0.6B-v2, Canary-1B-Flash, and Granite-4.0-1B-Speech models. Each gray line represents the angle $\phi(h_l, h_{l+1})$ as a function of layer $l$ for a single audio sample, while the dark line shows the averaged angle across a set of calibration speech samples.}
  \label{fig:angular}
\end{figure}

In addition to the layer-wise similarity patterns revealed by the CKA matrix in Figure \ref{fig:cka}, we examine complementary local representational dynamics across depth by measuring angular updates between adjacent layers. These updates are computed as the cosine angle between mean-centered activations, as illustrated in Figure \ref{fig:angular}. The resulting local patterns are consistent with the global CKA structure, indicating that many consecutive layers perform closely related transformations.

\subsection{Teacher-Student Representation Trajectories}
\label{appendix:trajectory}
To better understand why distilled block-recurrent student models retain high recognition accuracy despite aggressive parameter sharing, we visualize how internal representations evolve across encoder depth by comparing representation trajectories between teacher and student models. Specifically, we collect layer-wise encoder activations from a held-out evaluation set and project them into a two-dimensional subspace using principal component analysis (PCA).

\begin{figure*}[h]
\centering
\includegraphics[width=0.22\textwidth]{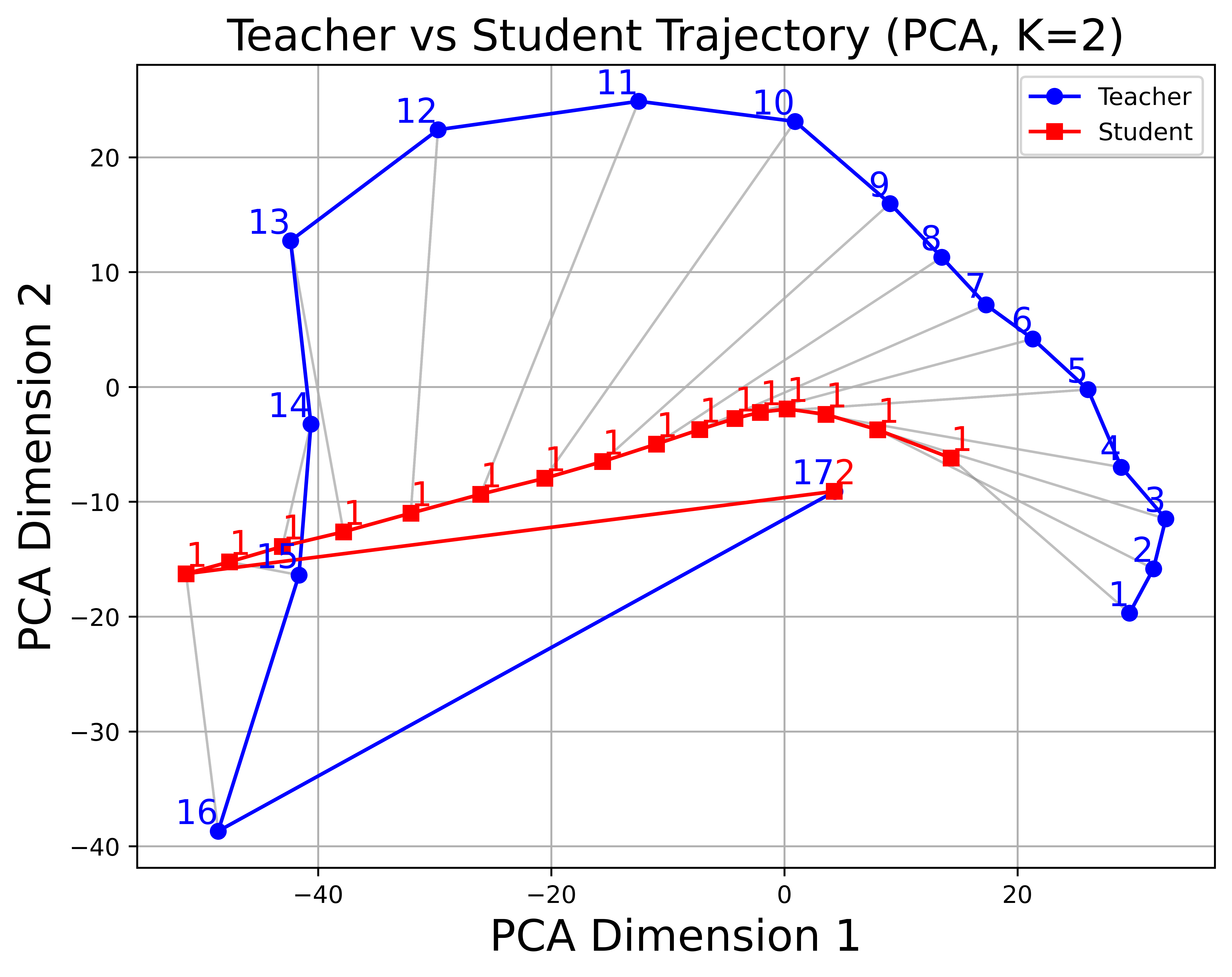}
\includegraphics[width=0.22\textwidth]{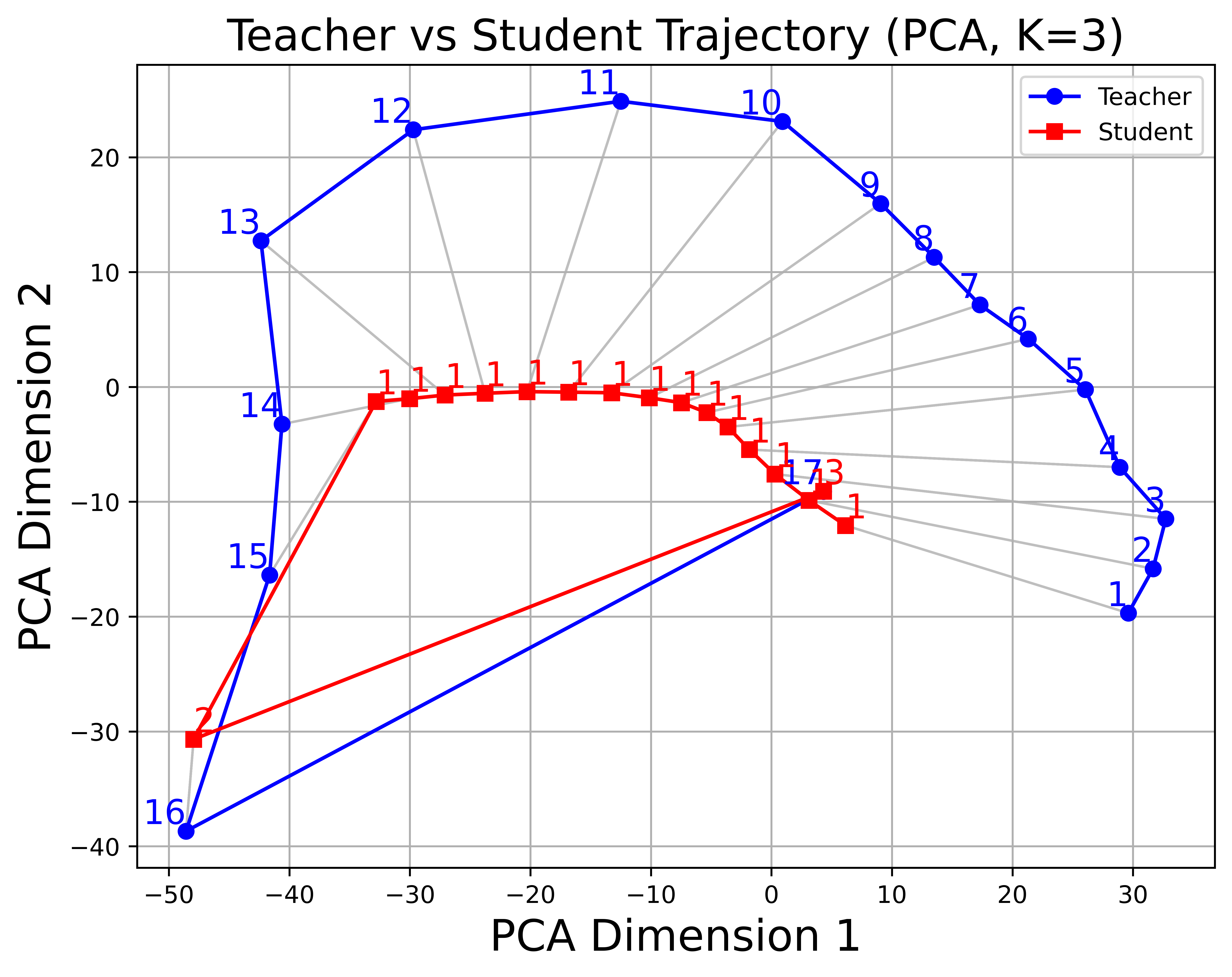}
\includegraphics[width=0.22\textwidth]{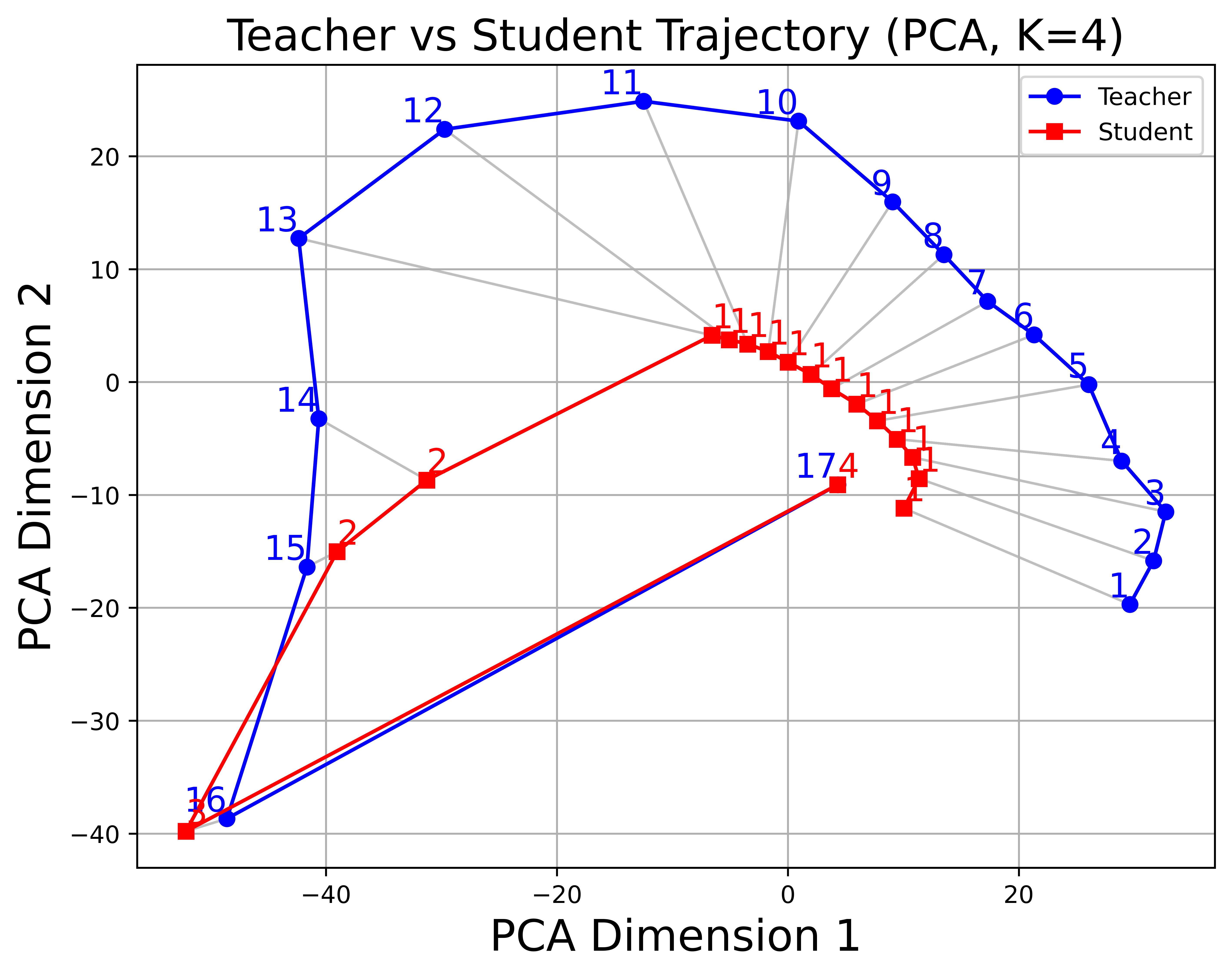}
\includegraphics[width=0.22\textwidth]{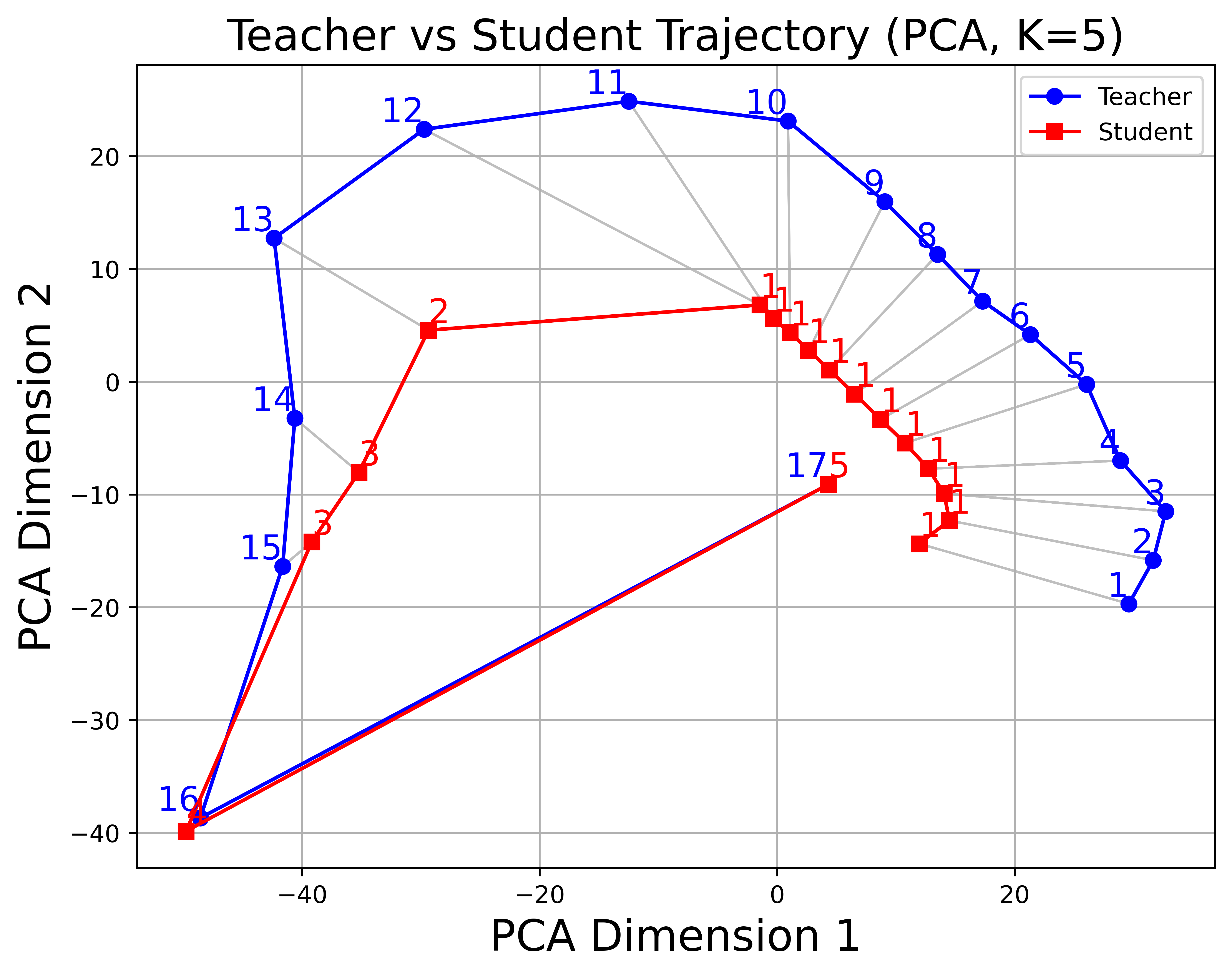}
\caption{Teacher–student projected representation trajectories under different block counts $K$. Blue curves denote the teacher encoder trajectory, while red curves show the block-recurrent student trajectory obtained via recurrent unrolling. Numerical annotations indicate layer indices for the teacher and block indices for the student.}
\label{fig:pca_trajectories}
\end{figure*}

Figure~\ref{fig:pca_trajectories} illustrates an example of projected trajectories for the teacher encoder (blue) and block-recurrent student encoders (red) for the same audio input under different block counts $K$. Across all values of $K$, the student trajectories follow a similar global progression to the teacher trajectory, despite using far fewer unique parameters. Through repeated application of each shared block, the student encoder performs iterative refinement that approximates the teacher’s intermediate representations along depth, demonstrating that block recurrence captures the teacher’s progressive computation. As $K$ increases, the student trajectory aligns more closely with the teacher trajectory, consistent with the improved recognition accuracy observed in Table \ref{tab:leaderboard_results}. These results provide visual evidence that block-recurrent encoders preserve the functional behavior of deep ASR encoders by reusing a small set of transformations.

\section{Additional Positioning Baselines}
\label{appendix:additional_baseline}

\subsection{Cross-model-family comparisons}
\label{appendix:baseline_cross}

The Open ASR Leaderboard serves as a useful reference for positioning our models relative to independently developed ASR systems. To further contextualize performance at comparable model scales, we compare against representative small and medium scale models reported on the leaderboard, including the widely adopted Whisper-large-v3-turbo and the Moonshine family of ASR models.

\begin{table}[H]
\centering
\caption{Cross-model-family positioning comparison}
\resizebox{0.5\linewidth}{!}{%
\begin{tabular}{c c c}
\hline
\textbf{Model} & \textbf{Params (M)} & \textbf{Avg. WER} \\
\hline
Moonshine-tiny & 30	& 12.54\% \\
Moonshine-streaming-tiny & 30	& 12.03\% \\
Edge-STT K=4 (ours) & 26.8	& 10.72\% \\
\hline
Moonshine-base & 60 & 9.67\% \\
Edge-STT K=6 (ours) & 39.5	& 9.65\% \\
\hline
Whisper-large-v3-turbo & 800	& 7.80\% \\
Edge-Parakeet K=4 (ours) & 105	& 7.62\% \\
\hline
\end{tabular}}
\label{tab:baseline_cross}
\end{table}

These results show that our compressed students remain competitive with independently developed ASR systems at similar parameter scales. We emphasize that this is intended as a positioning comparison rather than a controlled baseline, since the models differ substantially in architecture, training data, supervision scale, and optimization pipeline. Nevertheless, the results demonstrate that block-recurrent compression can produce compact ASR models whose accuracy is comparable to state-of-the-art systems trained independently from our framework.

\subsection{Memory-budget-matched quantization baseline}
\label{appendix:baseline_memory}

To further provide a baseline at comparable memory budgets within the model family, we apply a simple post-training weight quantization to the teacher encoders while keeping the original architecture unchanged. Specifically, we quantize encoder weights using round-to-nearest (RTN) quantization on linear and convolution layers. Results are shown below (W4/W3 denote encoder-only INT4/INT3 weight quantization, decoders are kept in fp16 precision):

\begin{table}[H]
\centering
\caption{Memory-budget-matched quantization baseline}
\resizebox{\linewidth}{!}{%
\begin{tabular}{c c c c c c c c c c c}
\hline
\textbf{Model} & \textbf{Effective memory (MB)} & \textbf{AMI} & \textbf{Earn}& \textbf{Giga}& \textbf{LS-c}& \textbf{LS-o}& \textbf{SPGI}& \textbf{TED}& \textbf{VoxP}& \textbf{Avg. WER}\\
\hline
STT Teacher & 217.8 & 26.23 & 20.56 & 14.15 & 2.54 & 5.97 & 5.80 & 5.58 & 7.60 & 11.05 \\
STT Teacher (W4) & 54.5 & 28.97 & 21.85 & 15.54 & 2.94 &7.37 & 6.77 & 6.33 & 8.32 & 12.26 \\
Edge-STT K=4 & 53.6  & 18.16 &  15.24 & 15.60 & 4.29 & 10.55 & 4.72 & 7.34 & 9.89 & 10.72 \\
\hline
Parakeet Teacher & 1217.8 & 11.16 & 11.09 & 9.73 & 1.70 & 3.19 & 2.18 & 3.37 & 5.96 & 6.05 \\
Parakeet Teacher (W4) & 304.5 & 12.53 & 11.55 & 9.97 & 1.76 &3.67 & 2.5& 3.45& 6.34 & 6.47 \\
Edge-Parakeet K=6 & 320.0 & 12.60 & 11.52 & 11.96 & 2.57 & 5.96 & 2.56 & 4.47 & 7.08 & 7.34 \\
Parakeet Teacher (W3) & 228.3 & 87.16 & 97.20 & 81.86 & 58.02 & 88.64 & 91.69 & 70.32 & 83.02  & 82.24 \\
Edge-Parakeet K=4 & 210.0 & 13.28 & 11.93 & 12.21 & 2.66 & 6.40 & 2.68 & 4.60 & 7.16 & 7.62 \\
\hline
\end{tabular}}
\label{tab:baseline_memory}
\end{table}

The results illustrate complementary trade-offs between structural compression and quantization. Importantly, quantization and structural compression are orthogonal techniques. Quantization reduces the precision of model weights, whereas block-recurrent compression reduces the number of unique parameters through weight sharing. Consequently, the compressed student models can themselves be quantized post-training or quantized-awared training, enabling memory footprints smaller than those achievable by either technique individually. 

\subsection{KL divergence baseline}
\label{appendix:baseline_loss}

Our student training objective combines representation-aligned encoder distillation with decoder-level KL distillation. To quantify the contribution of each component in Equation~\ref{eq:loss}, we perform a controlled ablation study on the Edge-STT model (K=4) and evaluate performance on the LibriSpeech test-clean and test-other splits (LS-c/LS-o).

\begin{table}[H]
\centering
\caption{KL divergence baseline}
\resizebox{0.7\linewidth}{!}{%
\begin{tabular}{c c c c}
\hline
\textbf{Training Objective} & \textbf{LS-c} & \textbf{LS-o} & \textbf{Avg. WER}\\
\hline
Decoder KL only & 4.65 & 11.65 & 8.15 \\
Task loss + decoder KL & 3.87 & 10.17 & 7.02 \\
Task loss + encoder KD + decoder KL (ours) & 3.84 & 9.79 & 6.81 \\
\hline
\end{tabular}}
\label{tab:baseline_loss}
\end{table}
Under aggressive compression, decoder-level KL distillation alone is insufficient to preserve model performance. Incorporating task supervision yields substantial gains, while the proposed encoder-level distillation provides an additional improvement beyond task supervision.

\subsection{Conventional LoRA fine-tuning baseline}
\label{appendix:baseline_lora}

For personalization, we additionally compare against standard LoRA fine-tuning using the same adaptation parameter budget ($5$ LoRA modules with rank $32$) but without user embeddings. This controlled comparison isolates the contribution of user-conditioned routing in MoE-LoRA, allowing us to determine whether the observed gains stem from speaker-specific conditioning rather than increased adaptation capacity. Following the setup in Figure \ref{fig:per_user_wer_comparison_stt_group8_rank32}, all experiments are conducted on the Edge-STT model with $K=8$ and evaluated on the VoxCeleb2 test set.

\begin{table}[H]
\centering
\caption{Conventional LoRA fine-tuning baseline}
\resizebox{0.4\linewidth}{!}{%
\begin{tabular}{c c}
\hline
\textbf{Model} & \textbf{Avg. WER (\%) }\\
\hline
LoRA baseline & 8.69 \\
MoE-LoRA (ours) & 8.03 \\
\hline
\end{tabular}}
\label{tab:baseline_lora}
\end{table}
MoE-LoRA reduces WER from $8.69\%$ to $8.03\%$. Given same adaptation parameter budget, this gain indicates that speaker-specific conditioning contributes additional personalization benefits beyond those achievable with standard LoRA alone.

\section{Latency and Energy Profiling Details}
\label{appendix:latency}
This appendix describes the experimental setup and implementation details used for latency and energy profiling of block-recurrent ASR models. All experiments use fixed batch size, sequence length, and precision settings across configurations. Reported latency and energy values are averaged across repeated runs, and relative comparisons are made against the STT-480ms reference model ($L=17$) to demonstrate the effect of block-recurrent weight sharing.

\textbf{Model Configurations} All latency experiments are conducted on the STT‑480ms encoder with an effective depth of $L=17$ layers and an input sequence length of $T=77$ frames. Block-recurrent configurations vary the number of shared blocks $K$, where each block is applied recurrently according to Table \ref{tab:grouping} to preserve the effective depth. 

\textbf{GPU Inference} GPU measurements are performed using custom Triton kernels tailored to the block-recurrent execution pattern. Two kernel variants are evaluated: (i) a scalar kernel that computes per-token outer products and operates primarily in a memory-bound regime, and (ii) a dot kernel based on \texttt{tl.dot}, which uses tensor-core-accelerated tiled matrix multiplication and is compute-bound. To encourage reuse of shared weights across recurrent applications, Triton cache-eviction hints (\texttt{evict\_last}) are enabled to promote L2 cache residency. Latency is measured using end-to-end kernel execution time averaged over multiple runs after warm-up.

\textbf{CPU Inference} CPU measurements use native PyTorch implementation without custom kernel fusion. Experiments are conducted on a server-class processor and Apple M4 Pro. Latency is reported as the average forward-pass time over multiple iterations. 

\textbf{Energy Measurement.}
GPU energy is measured via \texttt{nvmlDeviceGetTotalEnergyConsumption},
a cumulative hardware energy counter available on NVIDIA Ampere and later
architectures. The GPU is primed for the full measurement duration prior to
recording to reach thermal steady-state, then the counter is read before and
after a sustained pass of ${\approx}1000$ iterations. Average power is derived
as $P = \Delta E / \Delta t$, and per-inference energy is $E = P \times
t_{\text{infer}}$, where $t_{\text{infer}}$ is the measured latency by the interface. For AMD CPU, energy is measured via the Linux RAPL powercap interface (\texttt{/sys/class/powercap/intel-rapl}), which provides a direct hardware energy accumulator. RAPL is read before and after a timed window and then average power is derived and multiplied by per-inference latency. For Apple M4 Pro, energy is measured in the same way with the \texttt{zeus-apple-silicon} library. In Table~\ref{tab:dram_energy_grouping}, we report DRAM-only energy consumption to isolate memory traffic and quantify the impact of block-recurrent weight sharing on DRAM accesses and associated energy. DRAM energy consumption decreases consistently with stronger weight sharing. Compared to the full model (32.11 mJ), the block-recurrent models achieve up to a 1.46× reduction (21.98 mJ at $K{=}1$), with gains gradually diminishing as $K$ increases. Since memory accesses are substantially more energy-intensive than arithmetic operations \cite{Energy}, reducing memory traffic can yield significant energy savings even when FLOPs remain largely unchanged. The observed reduction in DRAM energy therefore suggests that block-recurrent weight sharing effectively decreases DRAM accesses and the associated memory-related energy overhead.



\begin{table*}[h]
\centering
\caption{%
  Energy per inference under weight sharing
  ($L{=}17$, $B{=}1$, $T{=}77$ frames).
  GPU energy via \texttt{nvmlDeviceGetTotalEnergyConsumption} (hardware counter,
  fp16); CPU energy via Linux RAPL (fp32). All values in Joules.
  Savings ratio = Full Model energy / K-group energy ($>{1}$ means less energy).
  Single run; energy std not independently measured (estimated ${<}2\%$).
  GPU: NVIDIA RTX PRO 6000 Blackwell Max-Q Workstation Edition (128\,MB L2).
  CPU: AMD Ryzen Threadripper PRO 7995WX (96 cores, 384\,MB L3).
}
\label{tab:energy_grouping}
\setlength{\tabcolsep}{4pt}
\renewcommand{\arraystretch}{1.1}
\begin{tabular}{cc cc}
\toprule
& & \textbf{GPU Scalar} & \textbf{CPU (fp32)} \\
\cmidrule(lr){3-3} \cmidrule(lr){4-4}
\textbf{Model} & $K$
  & Blackwell RTX Pro 6000
  & Threadripper 7995WX \\
\midrule
STT-480ms & \textendash
  & $2.019\ (1.00\times)$
  & $19.45\ (1.00\times)$ \\
\midrule
\multirow{5}{*}{Edge-STT}
 & 1 & $1.574\ \mathbf{(1.28\times)}$ & $18.17\ \mathbf{(1.07\times)}$ \\
 & 2 & $1.608\ (1.26\times)$           & $18.18\ (1.07\times)$ \\
 & 4 & $1.635\ (1.23\times)$           & $18.46\ (1.05\times)$ \\
 & 6 & $1.623\ (1.24\times)$           & $18.49\ (1.05\times)$ \\
 & 8 & $1.655\ (1.22\times)$           & $18.90\ (1.03\times)$ \\
\bottomrule
\end{tabular}
\end{table*}

\begin{table*}[h]
\centering
\caption{%
  DRAM energy per inference under weight sharing
  ($L{=}17$, $B{=}1$, $T{=}77$ frames, fp32) on Apple M4 Pro. All values in mili-Joules. Savings ratio = Full Model energy / K-group energy ($>{1}$ means less energy).
  }

\label{tab:dram_energy_grouping}
\setlength{\tabcolsep}{4pt}
\renewcommand{\arraystretch}{1.1}
\begin{tabular}{cc c}
\toprule
\textbf{Model} & $K$
  & Apple M4 Pro\\
\midrule
STT-480ms & \textendash
  & $32.11\ (1.00\times)$\\
\midrule
\multirow{5}{*}{Edge-STT}
 & 1 & $21.98\ \mathbf{(1.46\times)}$\\
 & 2 & $23.44\ (1.37\times)$\\
 & 4 & $24.38\ (1.31\times)$\\
 & 6 & $25.82\ (1.24\times)$\\
 & 8 & $27.34\ (1.17\times)$\\
\bottomrule
\end{tabular}
\end{table*}

\clearpage

\end{document}